%% file: icml_main.tex
\documentclass{article}

\usepackage{hyperref}

\usepackage[accepted]{icml2026}

\usepackage[utf8]{inputenc} %
\usepackage[T1]{fontenc}    %
\usepackage{url}            %
\usepackage{booktabs}       %
\usepackage{amsfonts}       %
\usepackage{nicefrac}       %
\usepackage{microtype}      %
\usepackage{xcolor}         %

\usepackage{graphicx}
\usepackage{multirow}
\usepackage{amsmath}
\usepackage{amssymb}
\usepackage{float}
\usepackage{algorithm}
\usepackage{orcidlink}
\usepackage{caption}        %
\usepackage{subcaption}
\usepackage{enumitem}       %
\input{preamble}

\icmltitlerunning{Analysis-by-Proxy: Localization Signals in VLM Condition Encoders}

\begin{document}

\twocolumn[
\icmltitle{Analysis-by-Proxy: Localization Signals in VLMs \\ Operating as Condition Encoders}

\begin{icmlauthorlist}
\icmlauthor{Yoav Baron}{tau}
\icmlauthor{Sara Dorfman}{tau}
\icmlauthor{Roni Paiss}{gdm}
\icmlauthor{Daniel Cohen-Or}{tau}
\icmlauthor{Or Patashnik}{tau}
\end{icmlauthorlist}

\icmlaffiliation{tau}{Tel Aviv University, Tel Aviv, Israel}
\icmlaffiliation{gdm}{Google DeepMind}

\icmlcorrespondingauthor{Yoav Baron}{yvbrn13@gmail.com}

\icmlkeywords{Vision-Language Models, Mechanistic Interpretability, Diffusion, Image Editing, Localization}

\vskip 0.3in
]

\printAffiliationsAndNotice{}

\input{sections/0_abstract_icml}

\input{figures/teaser_new_icml}

\input{sections/1_introduction_icml}
\input{sections/2_related_work_icml}

\input{sections/3_preliminaries_icml}

\input{sections/4_analysis_clean_icml}

\input{sections/5_clean_improving_icml}

\input{sections/7_conclusion_icml}

\section*{Acknowledgements}
The authors thank Amos Rottenberg and Alon Porat for their close support, friendship, and assistance in bringing this work to completion.
We also thank Daniel Garibi, Andrey Voynov and Omer Dahary for their early feedback and helpful suggestions. 
This work was supported in part by the Blavatnik Computer Science Research Fund.
The authors declare no competing financial interests.

\section*{Impact Statement}
This work advances the understanding of how Vision-Language Models (VLMs) encode
spatial information when run as part of an image editing pipeline, and leverages
these insights to improve the precision of text-guided image editing.
On the positive side, our Analysis-by-Proxy framework contributes to the growing
field of VLM interpretability, offering a transparent method to examine the internal
mechanisms of multimodal encoders.
Practically, providing users with more reliable and precisely localized editing tools
lowers the barrier to entry for creative professionals and everyday users.

However, we acknowledge the inherent dual-use risks associated with improvements in
generative editing capabilities.
Enhancing the spatial accuracy and structural preservation of image editing models
makes it easier to seamlessly modify visual content, which could be misused to
generate deepfakes, manipulate imagery, or spread disinformation.
While our research focuses on architectural analysis and foundational understanding
of these pipelines, the resulting techniques could be exploited maliciously.
Mitigating these societal risks will require continued investment in parallel
defenses such as robust watermarking, image provenance standards, and manipulation
detection systems — areas where deeper architectural interpretability, like the
insights provided in this work, may also prove beneficial.

\clearpage
\bibliographystyle{icml2026}
\bibliography{main}

\newpage
\appendix
\onecolumn
\input{sections/8_supp_icml}

\end{document}

%% file: preamble.tex
\usepackage{wrapfig}

\usepackage{tcolorbox}

%% file: sections/0_abstract_icml.tex
\begin{abstract}
Vision-Language Models (VLMs) are increasingly utilized as the conditioning backbone for diffusion-based image editing due to their remarkable multimodal reasoning capabilities.
While standalone VLMs demonstrate strong localization capabilities, editing pipelines frequently struggle to maintain this accuracy, particularly in complex, multi-entity scenes.
In this work, we investigate this performance gap, hypothesizing that it stems from treating the VLM as a \emph{condition encoder}.
In this role, the model is restricted to a single forward pass, preventing the autoregressive generation process for which it was optimized, thereby failing to fully expose its capabilities.
To investigate whether this spatial understanding persists when the VLM is used as a condition encoder, we introduce \textit{Analysis-by-Proxy}. In this framework, we train a lightweight, interpretable \textit{proxy} model on the VLM's intermediate representations using an auxiliary localization task.
By analyzing the VLM through this proxy, we uncover the specific VLM representations that encode localization information.
Our findings expose a fundamental mismatch between how spatial knowledge is represented within a VLM condition encoder and how it is extracted by current editing pipelines.
We reveal that under single-pass constraints, the localization signal does not reliably propagate to the predefined layer configurations commonly used for conditioning.
Instead, this crucial signal remains hidden within intermediate representations, at locations that vary depending on the input prompt.
Using our introduced \textit{Analysis-by-Proxy} framework, we reveal the fundamental failures of existing condition extraction strategies in editing pipelines, opening the door to more principled design of conditioning architectures.
\end{abstract}

%% file: figures/teaser_new_icml.tex
\begin{figure*}[t]
    \centering
    \includegraphics[width=0.88\linewidth]{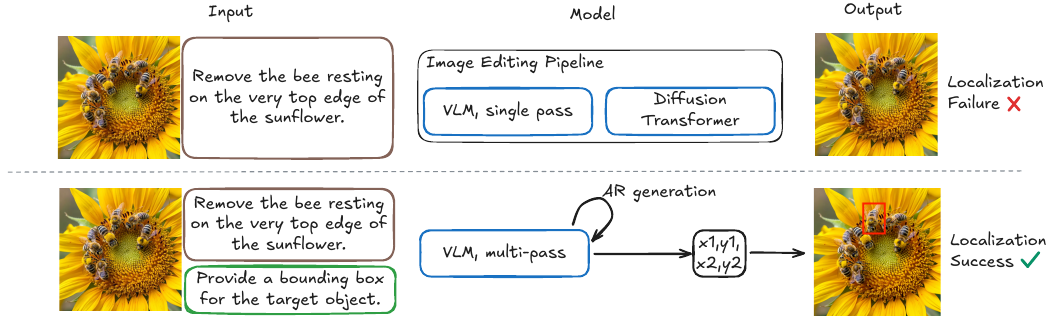}
    \caption{\small The intriguing discrepancy: standalone VLM localization succeeds, while VLM-conditioned editing mislocalizes the target.
    }
    \vspace{-10pt}
    \label{fig:teaser-new}
\end{figure*}

%% file: sections/1_introduction_icml.tex
\section{Introduction}
\label{sec:intro}

Vision-Language Models (VLMs)~\cite{bai2025qwen25vl, bai2025qwen3vl, liu2023llava, liu2023improvedllava, mistral2024pixtral, li2025surveystateartlarge} have recently emerged as powerful tools, demonstrating remarkable capabilities in parsing and reasoning over multimodal inputs.
As such, they have been widely adopted as the backbone for the instruction condition in state-of-the-art diffusion-based image editing models~\cite{wu2025qwen}. 
These editing pipelines typically condition a Diffusion Transformer (DiT)~\cite{rombach2022highresolutionimagesynthesislatent, esser2024scalingrectifiedflowtransformers} on the hidden representations extracted from a VLM, making the overall edit quality critically dependent on which internal representations are selected for conditioning.
These representations serve several functions within the editing process, including providing the signal for the accurate localization of the object or attribute to be edited.
Although localization is only one component of the editing process, even slight failures at this stage directly result in incorrect, misplaced, or entirely hallucinated edits.
The challenge of accurate localization is especially pronounced in complex, multi-entity scenes, where the model must determine which visual instance satisfies the textual description and distinguish it from similar surrounding objects
(see Figure~\ref{fig:motivation}).

In this work, we investigate the behavior of the VLM when it serves as the conditioning backbone for a DiT.
We characterize this paradigm as treating the VLM as a \emph{condition encoder}: the model processes the input in a \textbf{single} forward pass without autoregressively generating text.
In this setting, standard practice utilizes representations from a predefined and input-independent subset of layers for the conditioning signal.  
Regardless of the backbone's input modality,
standard procedure includes using only the final-layer tokens~\cite{labs2025flux1kontextflowmatching, wu2025qwen}, pooling hidden states across layers~\cite{hacohen2026ltx2efficientjointaudiovisual, wang2025comprehensivestudydecoderonlyllms}, or feeding features from different layers into corresponding layers of the DiT~\cite{flux-2-2025, liu2024playgroundv3improvingtexttoimage, gutflaish2025generatingimage1000words}.
While most editing pipelines rely on single-modality conditioning, recent architectures such as Qwen-Image-Edit~\cite{wu2025qwen} leverage a multimodal approach by providing both image and text inputs to the conditioning VLM.

Most existing methods for analyzing information flow in VLMs rely on the model autoregressively generating text ~\cite{kaduri2024_vision_of_vlms, cohen2026performancegapentityknowledge, nikankin2025taskdifferentcircuitsdisentangling}.
Consequently, despite the growing adoption of VLMs as condition encoders, their internal behavior in this restricted operating mode remains under-explored.
Our analysis reveals a striking performance gap: while the editing pipeline often fails to localize the intended target, the underlying VLM successfully identifies the correct object when allowed to autoregressively generate text (see Figure~\ref{fig:teaser-new}).
We hypothesize that this discrepancy is a consequence of the model's pre-training objective.
The VLM's internal representations are heavily optimized for an autoregressive generation paradigm.
As a result, when restricted to a \textit{single} forward pass, spatial knowledge that would typically emerge through sequential decoding does not necessarily fully propagate to the layers extracted for the condition.
Thus, while the model successfully encodes this spatial information internally, it may not be exposed in the conditioning signal provided to the DiT.
\input{figures/motivation_icml}

Our goal is to demonstrate that while this spatial information is diluted from the VLM's output when used as a condition encoder, it remains encoded within the VLM's internal representations.
Furthermore, we aim to develop the means to recover these hidden signals from the network's intermediate representations.
Directly probing the VLM for spatial knowledge under this single-pass restriction is challenging, as we must extract this information directly from the continuous hidden states without relying on autoregressive decoding.
To address this challenge, we introduce \textit{Analysis-by-Proxy}.
In this framework, we isolate spatial knowledge by training a lightweight, interpretable \textit{proxy} model on a dedicated auxiliary task.
By training this proxy on the VLM's internal representations, we can \textit{analyze} which layers and tokens are most significant to its performance, thereby uncovering the underlying information flow within the VLM.

Applying this framework to the condition encoder setting reveals that spatial information is distributed highly unevenly across the model's layers.
Crucially, representations in the final layer are notably poor at conveying spatial details, demonstrating that the common practice of using only these final hidden states is fundamentally limiting.
Furthermore, while intermediate layers contain much stronger spatial signals, the specific layers where these signals peak shift dynamically depending on the input.
Consequently, conditioning methods that rely on extracting features from any predefined configuration of layers are inherently suboptimal.

At the token level, we observe that the spatial signal is not uniformly spread across the input sequence. Instead, it is sparsely encoded and concentrated almost exclusively within a few specific tokens.
These dominant tokens strongly correspond to the semantically significant nouns and adjectives that define the target edit.
Furthermore, we demonstrate that utilizing our proxy's outputs enables improved edit localization in complex scenes.
Ultimately, our findings establish a deeper understanding of the internal mechanisms of VLMs when operating as condition encoders,
alongside a structured and flexible framework for analyzing models in this setting. This, in turn, provides a principled foundation for exploring other conditioning architectures within the design space of text-guided editing pipelines.

%% file: figures/motivation_icml.tex
\begin{figure*}[t]
    \centering
    \begin{minipage}{0.95\linewidth}
    \centering
    \begin{tabular}{@{} c@{\hspace{2pt}}c @{\hspace{6pt}} c@{\hspace{2pt}}c @{\hspace{6pt}} c@{\hspace{2pt}}c @{\hspace{6pt}} c@{\hspace{2pt}}c @{}}

        \multicolumn{2}{@{}c@{\hspace{6pt}}}{%
            \begin{minipage}[t]{0.23\linewidth}
                \centering \tiny ``Break the graphite tip off the pencil lying second from the right.''
            \end{minipage}%
        } &
        \multicolumn{2}{c@{\hspace{6pt}}}{%
            \begin{minipage}[t]{0.23\linewidth}
                \centering \tiny ``Change the paint of the third Vespa from the left to a pale mint green.''
            \end{minipage}%
        } &
        \multicolumn{2}{c@{\hspace{6pt}}}{%
            \begin{minipage}[t]{0.23\linewidth}
                \centering \tiny ``Change the balloon flying lowest in the sky to solid bright yellow.''
            \end{minipage}%
        } &
        \multicolumn{2}{c@{}}{%
            \begin{minipage}[t]{0.23\linewidth}
                \centering \tiny ``Remove the clothespin clipped furthest to the right.''
            \end{minipage}%
        } \\[10pt]

        \includegraphics[width=0.115\linewidth]{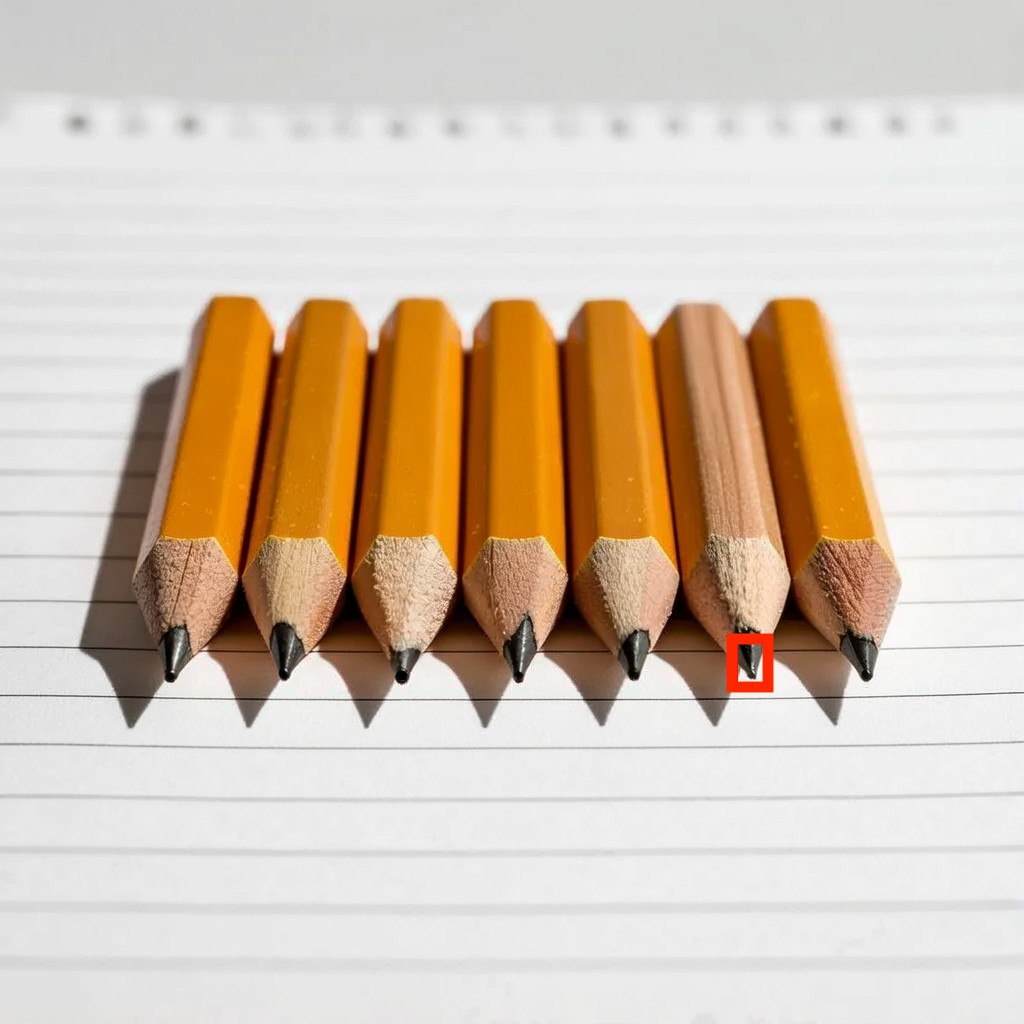} &
        \includegraphics[width=0.115\linewidth]{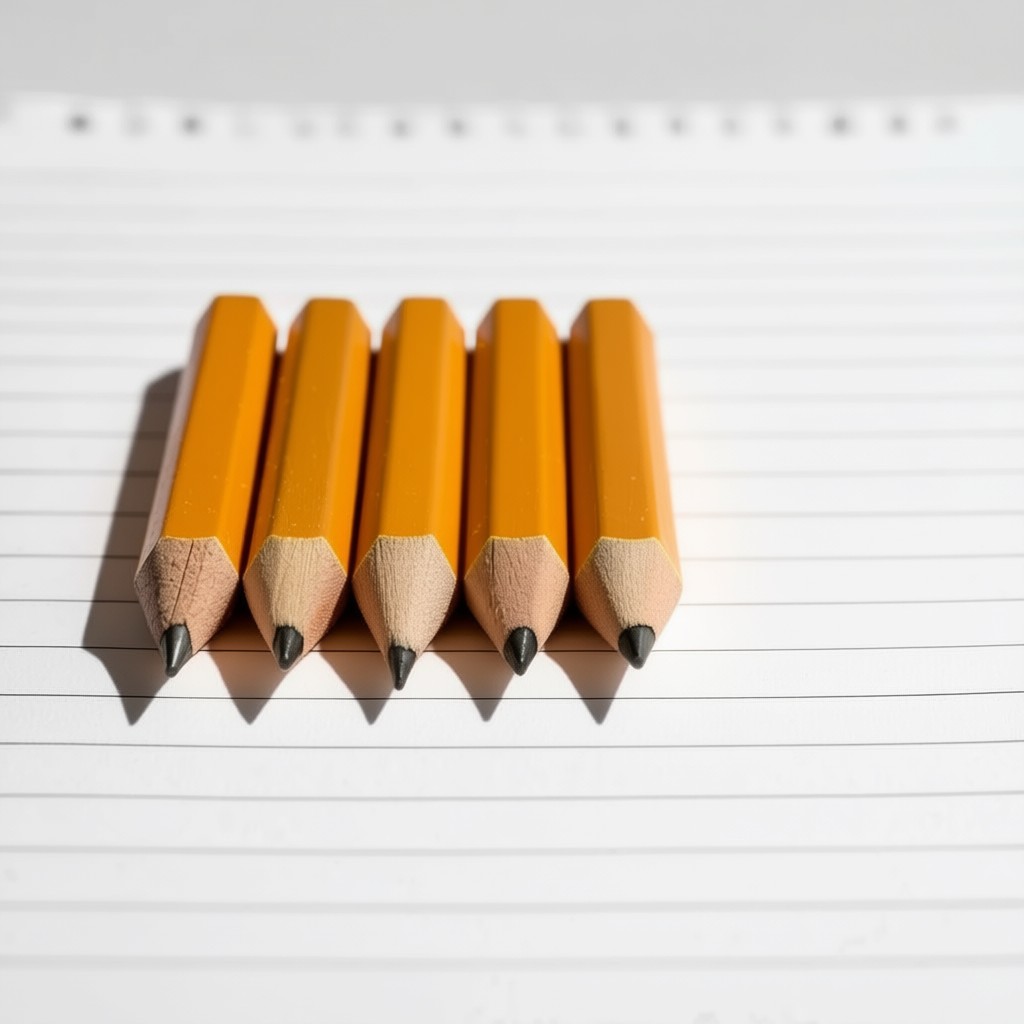} &

        \includegraphics[width=0.115\linewidth]{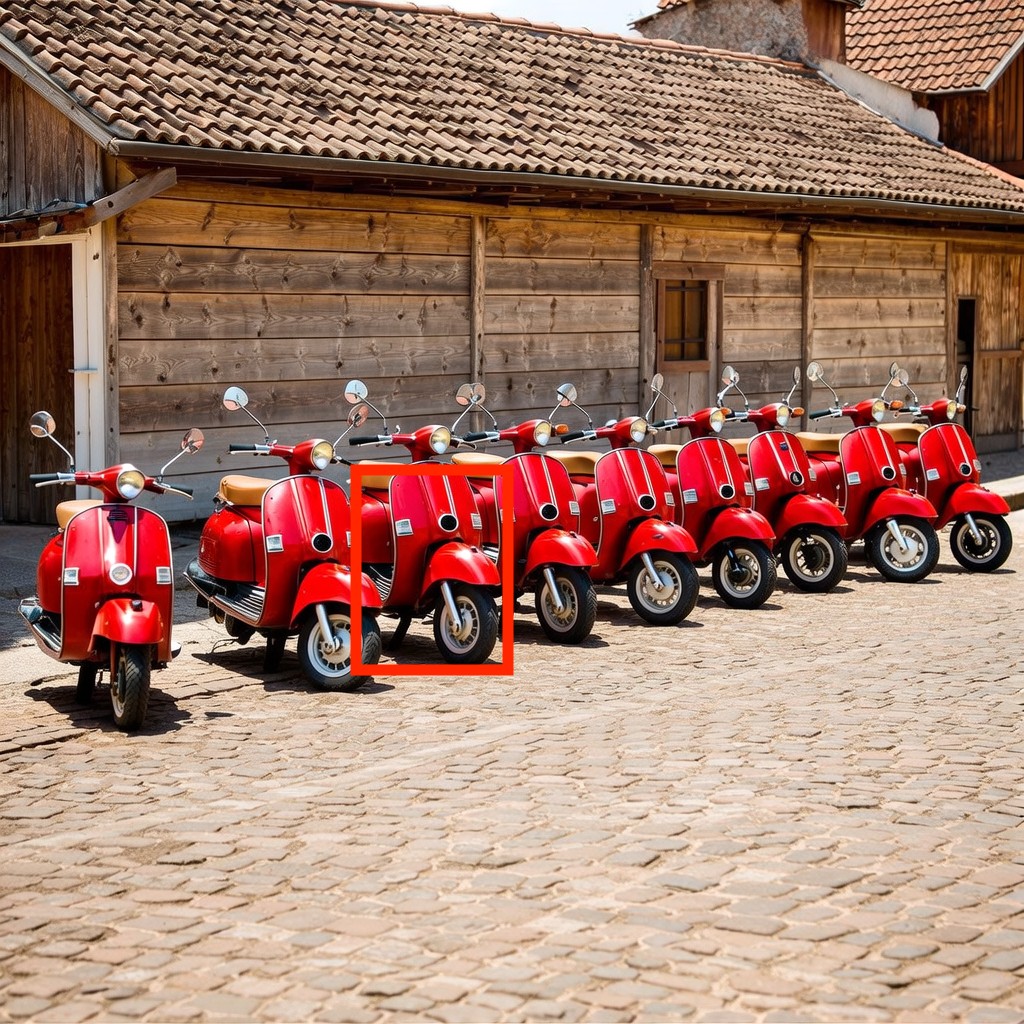} &
        \includegraphics[width=0.115\linewidth]{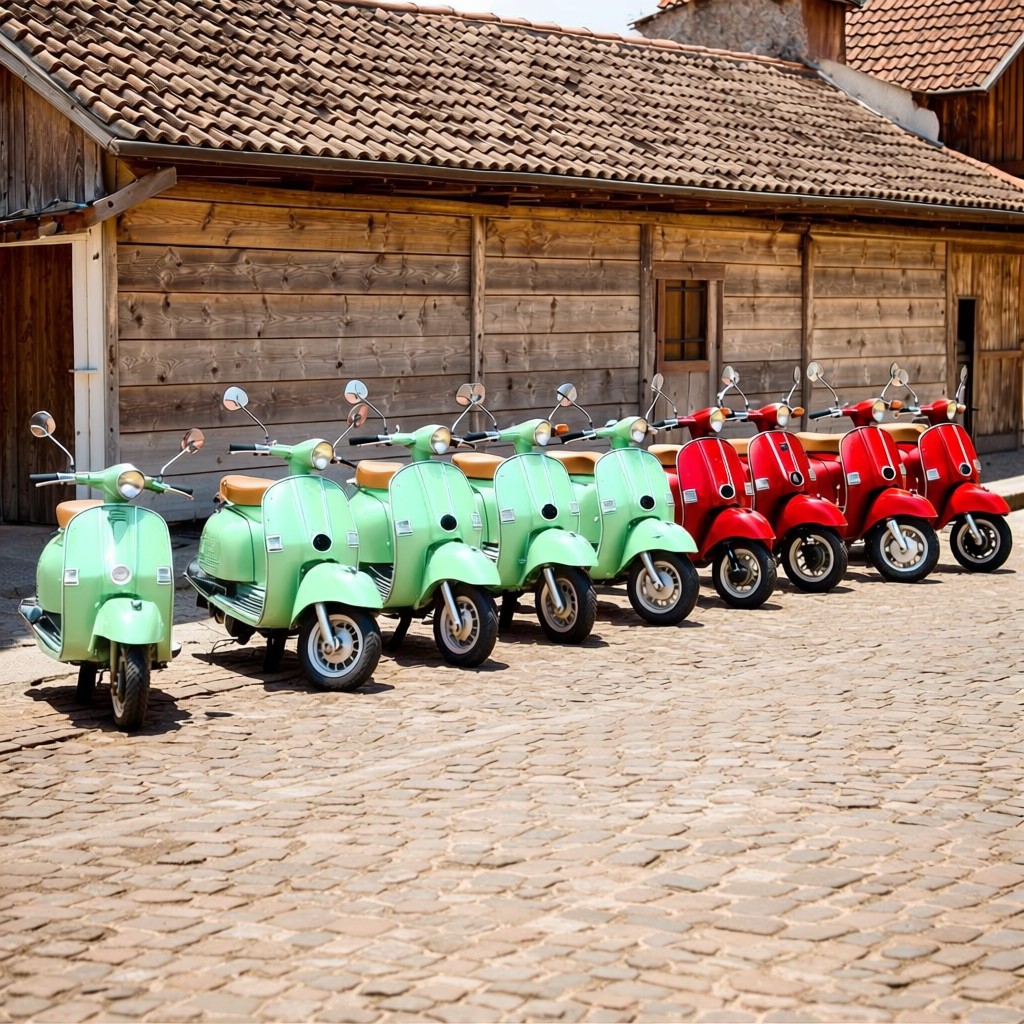} &

        \includegraphics[width=0.115\linewidth]{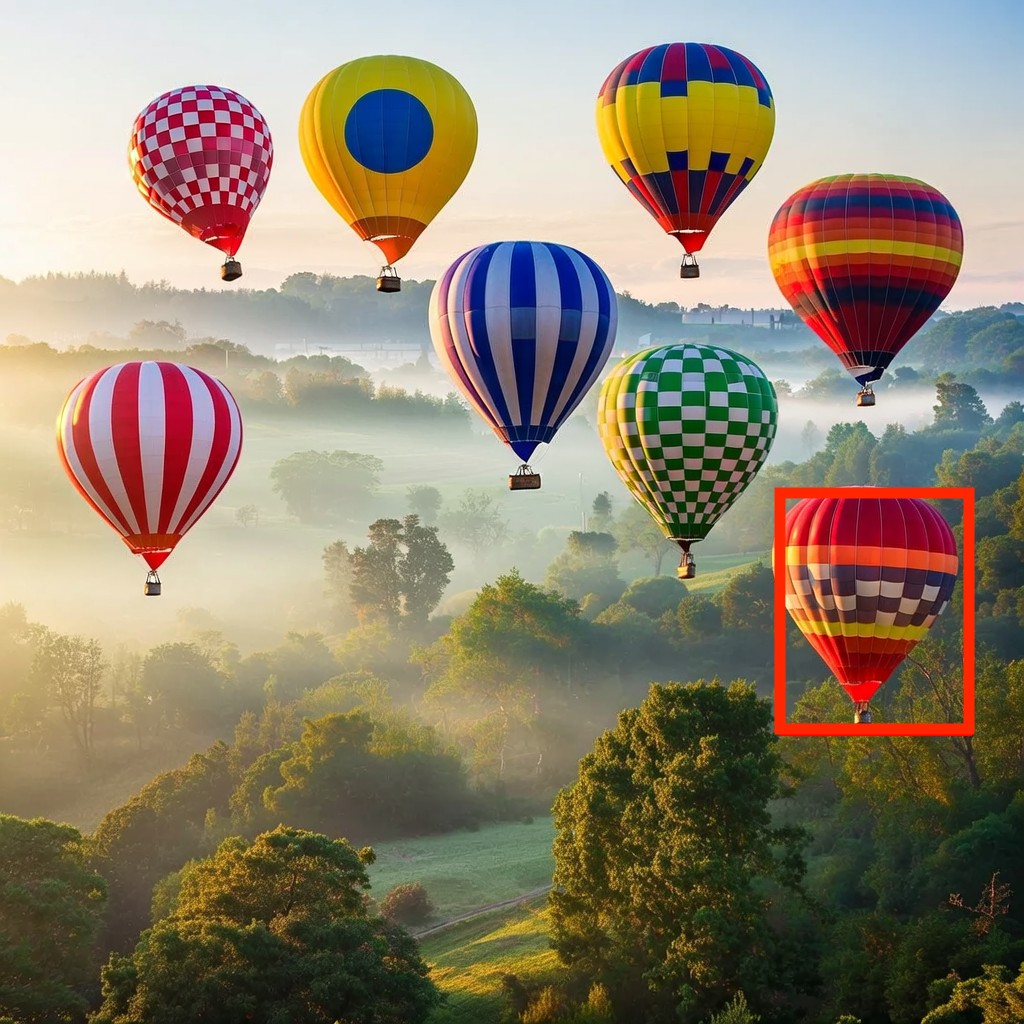} &
        \includegraphics[width=0.115\linewidth]{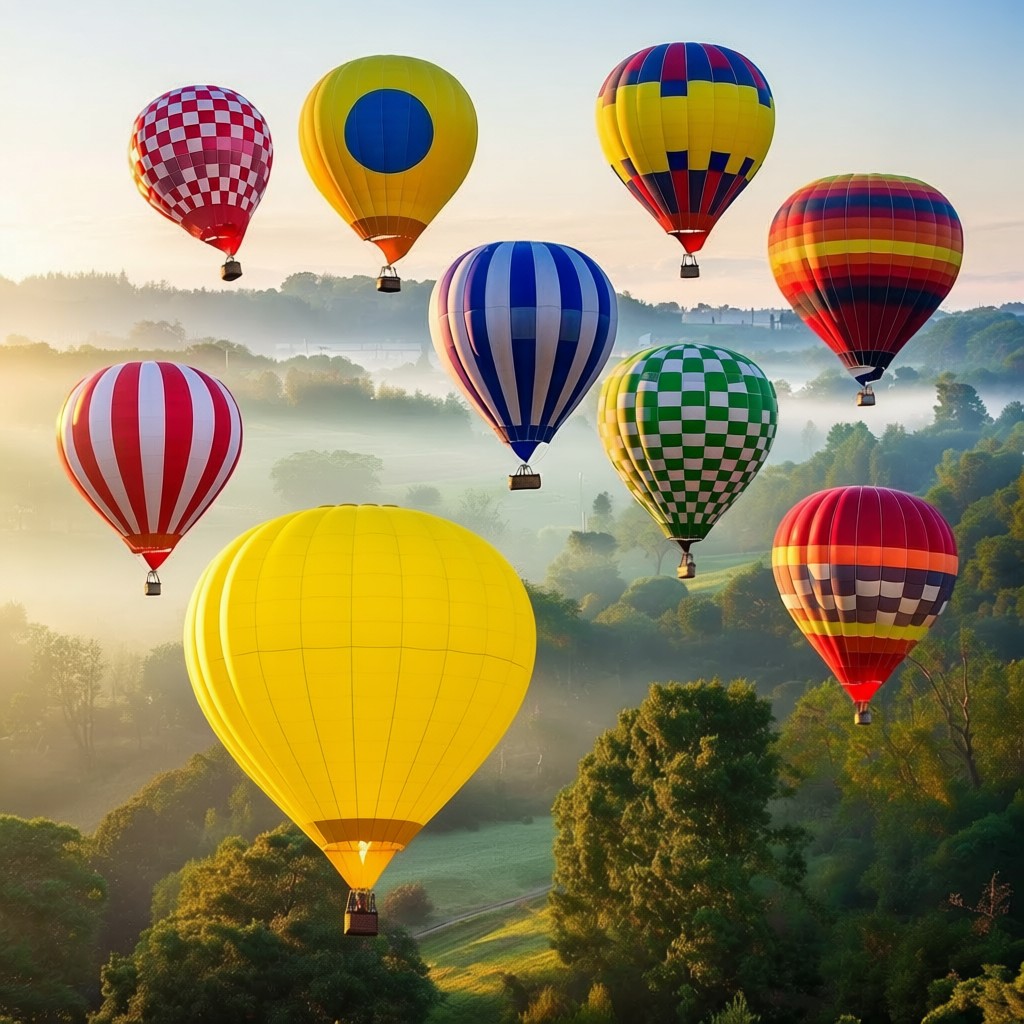} &

        \includegraphics[width=0.115\linewidth]{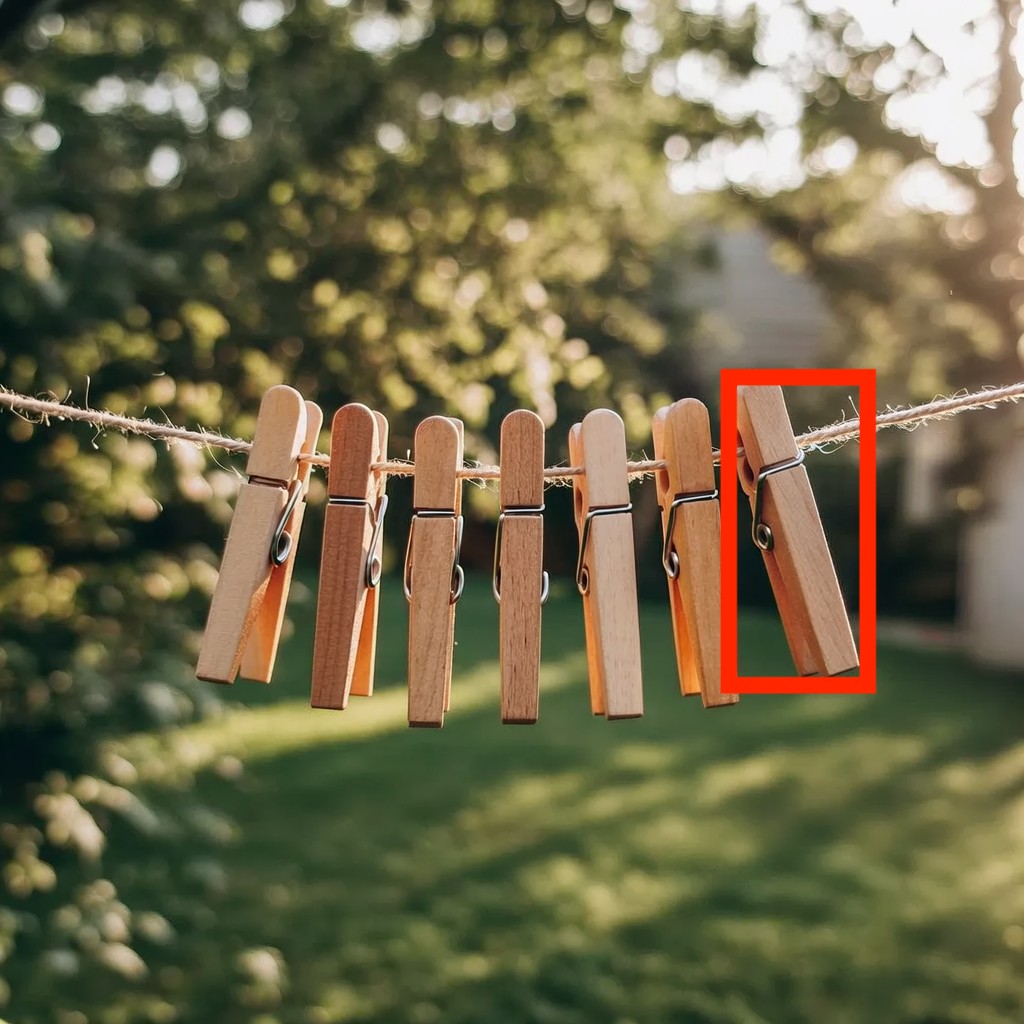} &
        \includegraphics[width=0.115\linewidth]{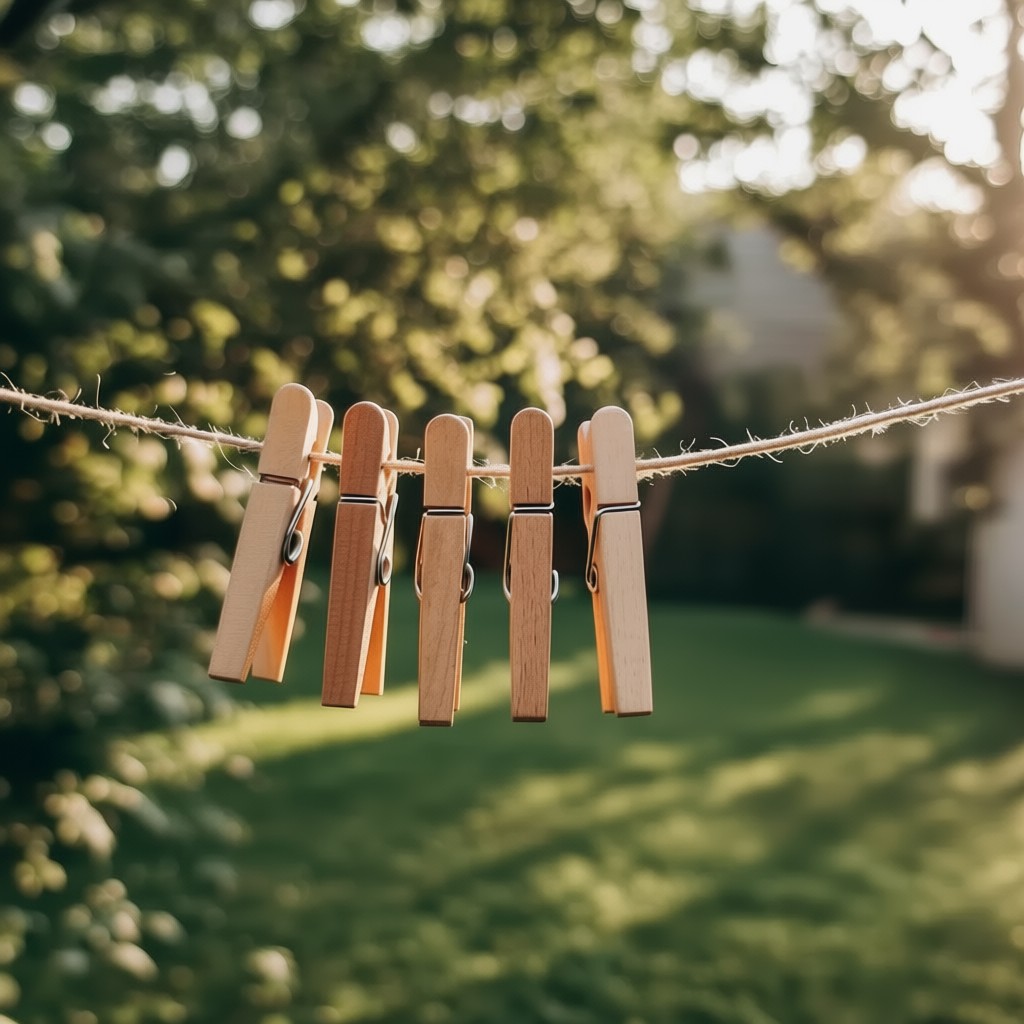} \\[-4pt]

        \fontsize{4}{5}\selectfont \begin{tabular}[t]{@{}c@{}}VLM prediction\end{tabular} & \fontsize{4}{5}\selectfont \begin{tabular}[t]{@{}c@{}}Localization\\failure\end{tabular} &
        \fontsize{4}{5}\selectfont \begin{tabular}[t]{@{}c@{}}VLM prediction\end{tabular} & \fontsize{4}{5}\selectfont \begin{tabular}[t]{@{}c@{}}Color\\leakage\end{tabular} &
        \fontsize{4}{5}\selectfont \begin{tabular}[t]{@{}c@{}}VLM prediction\end{tabular} & \fontsize{4}{5}\selectfont \begin{tabular}[t]{@{}c@{}}Hallucination\end{tabular} &
        \fontsize{4}{5}\selectfont \begin{tabular}[t]{@{}c@{}}VLM prediction\end{tabular} & \fontsize{4}{5}\selectfont \begin{tabular}[t]{@{}c@{}}Excessive\\removal\end{tabular} \\

    \end{tabular}
    \end{minipage}

    \caption{ Failure cases across diverse editing tasks. For each pair: \textbf{Left:} Inputs overlaid with accurately extracted VLM bounding boxes. \textbf{Right:} Resulting failed edits.
    From left to right, failure modes include: localization errors (modifying the wrong object in a sequence),  leakage (applying an intended attribute to multiple similar objects), hallucination (generating a new object rather than altering the target), and excessive removal. The accurate bounding boxes reveal the VLM's inherent potential to precisely distinguish what needs to be edited; unfortunately, the pipeline struggles to capitalize on this ability, resulting in failed execution.} %
    \vspace{-12pt}
    \label{fig:motivation}
\end{figure*}

%% file: sections/2_related_work_icml.tex
\section{Related Work}

\paragraph{\textbf{Vision-Language Models (VLMs).}}

Vision-Language Models (VLMs)~\cite{bai2025qwen3vl, bai2025qwen25vl, wang2025internvl35advancingopensourcemultimodal} are typically built upon pretrained LLMs and extended to process visual inputs. Images are passed through a vision encoder and projected into the LLM's input space using a lightweight adapter~\cite{merullo2023linearlymappingimagetext, tsimpoukelli2021multimodalfewshotlearningfrozen}. Through training for autoregressive text generation with a standard language modeling objective, these models learn to generate textual outputs grounded in the provided visual context.

\paragraph{\textbf{Text Conditioning in Image Editing Models.}}
Current image editing architectures~\cite{wu2025qwen, flux-2-2025, gutflaish2025generatingimage1000words} employ an LLM or VLM as a text backbone to encode edit instructions. The resulting representations are then used to condition a Diffusion Transformer (DiT)~\cite{rombach2022highresolutionimagesynthesislatent, esser2024scalingrectifiedflowtransformers} that generates the output image.
Different methods utilize distinct strategies for extracting these condition embeddings. FIBO~\cite{gutflaish2025generatingimage1000words} feeds features from various LLM layers into corresponding layers of the DiT, whereas Qwen-Image-Edit~\cite{wu2025qwen} extracts signals exclusively from the final layer of a VLM. Alternatively, FLUX.2~\cite{flux-2-2025} concatenates activations from layers 10, 20, and 30 across the channel dimension, but notably does not pass the input image to the VLM backbone.
In all such pipelines, the text backbone is deployed purely as a non-generative encoder restricted to a single forward pass. In this work, we analyze how VLMs operate under this single-pass regime and demonstrate that existing extraction strategies under-utilize the capabilities of the text backbone.

\paragraph{\textbf{VLM Interpretability}}

Following prominent methods for interpreting LLMs~\cite{dar2023analyzingtransformersembeddingspace, geva2021transformerfeedforwardlayerskeyvalue, geva2022transformerfeedforwardlayersbuild, nostalgebraist2020interpreting}, recent studies on VLM interpretability analyze internal mechanisms to better understand model predictions.
Most of these works~\cite{cohen2026performancegapentityknowledge, nikankin2025taskdifferentcircuitsdisentangling, kaduri2024_vision_of_vlms, neo2025interpretingvisualinformationprocessing, liu2025seeingbelievingprobingdisconnect} require sampling from the VLM in a standard autoregressive setting, interpreting the model based on its generated responses.
For instance, Cohen et al.~\cite{cohen2026performancegapentityknowledge} and Nikankin et al.~\cite{nikankin2025taskdifferentcircuitsdisentangling} pinpoint discrepancies in question-answering accuracy when the same prompt is conveyed via different modalities.
Kaduri et al.~\cite{kaduri2024_vision_of_vlms} and Neo et al.~\cite{neo2025interpretingvisualinformationprocessing} identify subject-level localization signals directly within the generated tokens.
Recently, Jiang et al.~\cite{jiang2024interpretingeditingvisionlanguagerepresentations} proposed a method that analyzes model representations directly, without relying on sampling, by applying Logit Lens~\cite{nostalgebraist2020interpreting} on the hidden representation on the VLM. Notably, their focus remains on the final prediction.
In contrast to these methods, our work aims to analyze the VLM as a component in an image editing pipeline, rather than in its natural function. In this setting, the VLM is not sampled, meaning that we cannot rely on it generating any tokens.

%% file: sections/3_preliminaries_icml.tex
\section{Preliminaries}

\paragraph{\textbf{Vision-Language Models (VLMs) and Hidden States.}}
Standard VLMs process an input image $I$ and a textual instruction $T$ to form a multimodal sequence of length $M$.
Concretely, Qwen2.5-VL consists of 28 transformer layers.
We denote the full sequence of hidden states at layer $l$ as $H^{(l)} \in \mathbb{R}^{M \times d}$, where $l \in \{0, \dots, 27\}$ and $d$ is the hidden dimension.
The hidden state of the $i$-th token at layer $l$ is denoted by $h_i^{(l)} \in \mathbb{R}^{d}$.
For a subset of layers $L \subseteq \{0,\dots,27\}$, we denote the corresponding hidden states by $\{H^{(l)}\}_{l \in L}$, and the representations of token $i$ across these layers by $\{h_i^{(l)}\}_{l \in L}$.
Furthermore, we denote the attention assigned by the query corresponding to token $i$ to the key corresponding to token $j$, averaged across all attention heads at layer $l$, as $\alpha^{(l)}_{i,j}$.

\paragraph{\textbf{Diffusion-Based Image Editing.}}
Modern image editing pipelines generate a modified image $\hat{I}$ from a source image $I$ and an instruction $T$ using a Multi-Modal Diffusion Transformer (MMDiT).
The diffusion model reverses a Gaussian noise process, conditioned on representations extracted from a VLM.
In such pipelines the VLM acts as a \emph{condition encoder}: $I, T$ are processed through a single forward pass, and a subset of the resulting hidden states is supplied to the diffusion model.

In our experiments we analyze the Qwen-Image-Edit pipeline, which uses the Qwen2.5-VL-7B model as its VLM backbone and conditions the diffusion model on the final-layer representations $H^{(27)}$.

\paragraph{\textbf{Q-Former and Proxy Formulation.}}
To probe the VLM, we employ a Q-Former as a proxy.
As depicted in Figure~\ref{fig:qformer_arch}, the Q-Former is a minimal Transformer with $N$ learnable queries, $Q \in \mathbb{R}^{N \times d}$.
Through multi-head attention, these queries cross-attend to a subset of VLM hidden states $\{H^{(l)}\}_{l \in L}$, while also attending to one another via self-attention.

The refined query representations are passed to a learned coordinate head, which maps these features to predicted bounding box coordinates $(x_1, y_1, x_2, y_2)$.

We index the attention heads using a single index $h$, where $h=(m,j)$ identifies the $j$-th head in layer $m$.
Let $a_{h,q}^{(l,i)}$ denote the cross-attention weight from proxy query $q$ in head $h$ to the $i$-th token of VLM layer $l$.

%% file: sections/4_analysis_clean_icml.tex
\section{On the Source of Localization Failures}

As demonstrated in Figure~\ref{fig:motivation}, existing image editing pipelines often fail on multi-entity scenes.
In this section, we analyze the causes of these failures by probing how well localization-relevant information is encoded across the layers of the VLM.
We conduct this analysis using a process we term \emph{Analysis-by-Proxy}, where we probe~\cite{belinkov2022probing, belinkov2019analysis} the VLM’s hidden representations through a lightweight, well-defined localization task.
In this analysis, we apply our framework to a prominent modern editing pipeline, Qwen-Image-Edit \cite{wu2025qwen}, and utilize its default Qwen2.5-VL-7B \cite{bai2025qwen25vl} backbone as our primary case study.

\subsection{The Localization Gap}
We begin our analysis by revealing that the suboptimal localization does not stem from an inherent knowledge gap in the VLM itself but from its setting within the editing pipeline.
For this, we conduct a baseline evaluation. We construct a curated evaluation set consisting of 200 complex, multi-entity scenes, paired with specific local editing instructions.

For each example, we evaluate localization accuracy across two distinct settings. First, we evaluate the end-to-end pipeline by assessing whether the downstream DiT correctly localizes the edit to the target object. Second, for the standalone VLM evaluation, we directly prompt the model to output the target's bounding box coordinates via autoregressive text generation.
The accuracy of both tasks is determined via human evaluation. An edit is considered successful if it alters only the target subject and nothing else, and a bounding box is considered accurate if it wholly encompasses only the target subject and nothing else.

Our evaluation highlights a clear performance gap. The VLM accurately predicts the target bounding box in $89.0\%$ of the samples, while the full pipeline successfully localizes the edit in only $57.5\%$.
This \textbf{31.5\%} drop presents an intriguing empirical discrepancy: while the VLM exhibits strong spatial reasoning capabilities, the downstream application utilizing it performs significantly worse.

This discrepancy may originate from two sources: (1) the DiT does not effectively leverage the spatial cues present in the VLM representations; or (2) the conditioning signal extracted from the VLM does not preserve localization information in a sufficiently decodable form.
In this work, we investigate the latter possibility.
In the localization experiment described above, the VLM operates autoregressively with an explicit localization objective, whereas in the editing pipeline the conditioning is typically obtained from a single forward pass, often using only the final-layer hidden states. This usage differs from the VLM’s autoregressive generation setting, and hence may fail to properly expose the spatial signal needed for localization.

We want to investigate whether the localization signals also exist in the VLM representations generated by a single forward pass only. We do so by probing the model’s internal hidden states.

Our goal is to determine whether the VLM’s demonstrated capacity for precise localization, typically elicited through explicit autoregressive prompting, can
also be recovered directly from the hidden states of a single forward pass. We show that this is indeed possible by leveraging the full set of hidden states within a single forward pass, without relying on unconstrained autoregressive generation or modifying the input prompt.
Crucially, we further demonstrate that distilling this recovered localization signal into a dedicated conditioning input is effective and significantly improves the editing model’s localization performance.

\input{figures/proxy_training_icml}
\subsection{Analysis-by-Proxy}

We introduce the \textit{Analysis-by-Proxy} framework (Figure~\ref{fig:proxy_training}) to (i) investigate whether precise localization signals exist within the VLM representations 
produced by a single forward pass,
and (ii) determine where in the model these signals are encoded in their most decodable form.

\noindent
The core principle of our approach is to employ a lightweight model that acts as a \textbf{proxy} for the downstream DiT in order to \textbf{analyze} the VLM representations. The proxy is trained on a tractable task that replaces the complex editing objective of the DiT. Specifically, it is trained to predict the explicit bounding box of a local edit directly from the VLM’s hidden states.

We utilize a Q-Former \cite{li2023blip2bootstrappinglanguageimagepretraining} as our proxy model, as it offers three key advantages over the DiT for analyzing the VLM hidden representations:
\begin{enumerate}[leftmargin=*]
    \item \textbf{Inherent Interpretability:} The Q-Former’s learned queries are explicitly supervised to predict localization, compelling them to extract spatial cues from the VLM hidden states. Unlike the DiT, where localization is implicitly entangled within the diffusion objective, these dedicated tokens enable us to trace how spatial information propagates from the VLM representations into explicit localization outputs.
    \vspace{16pt}
    \item \textbf{Architectural Simplicity:} Compared to the DiT, the Q-Former operates with substantially simpler mechanics. It requires only a single forward pass, rather than multiple denoising steps, consists of fewer and smaller components, and is significantly more lightweight to train.

    \item \textbf{Signal Clarity:} The Q-Former is trained on a single localization objective, resulting in substantially cleaner internal activations. In contrast to the DiT’s multi-objective generative representations, this setting reduces confounding factors and enables a more focused analysis.
\end{enumerate}

\subsection{Localization Signals in the VLM}

\input{figures/q-former_convergence_icml}

We first employ the proxy framework to identify if and where localization information is encoded within the VLM, when it is used as a non-generative condition encoder. Concretely, we train a series of independent Q-Formers (Figure \ref{fig:proxy_train_fig}). Initially, we train a separate proxy model for each VLM layer, providing the full sequence of hidden states $H^{(l)}$ from a single layer $l$ as input. Given the accelerated convergence observed in the intermediate layers, we subsequently train a specialized proxy conditioned exclusively on the hidden states of the user prompt tokens,
$\{h_i^{(l)}\}_{l\in L}$, from middle layers $L=\{15,\dots,24\}$.
All proxies are trained on a dataset of triplets consisting of an input image $I$, an editing prompt $T$, and the target bounding box coordinates $B$ natively predicted by the unconstrained VLM. Let $\hat{B}$ denote the spatial coordinates predicted by the proxy. 

We optimize the models using a mixed bounding box regression objective combining Generalized Intersection over Union (GIoU) \cite{Rezatofighi_2018_CVPR} and $L_1$ loss:
$$\mathcal{L}(B, \hat{B}) = \lambda_{\text{GIoU}} \mathcal{L}_{\text{GIoU}}(B, \hat{B}) + \lambda_{L_1} \mathcal{L}_{L_1}(B, \hat{B}),$$
where $\lambda_{\text{GIoU}}$ and $\lambda_{L_1}$ are hyperparameters balancing the two loss components.

Figure~\ref{fig:qformer_converge} presents the mean localization error (bounding box center distance) along the training process, illustrating the decodability of spatial signals across VLM layers.

The proxy converges significantly faster and achieves superior bounding box predictions when conditioned on intermediate layers rather than the final layer. This stark contrast indicates that final-layer representations over-abstract or entangle crucial spatial cues, making mid-layer extraction essential to fully realize the VLM's localization potential for precise editing.

\vspace{-1pt}
\subsection{Q-Former Decomposition}
\vspace{-1pt}
\label{lab:q-comp}
\input{figures/method_icml}

Since spatial decodability varies across the network, we next examine how the input context determines which layer holds the strongest signal~\cite{chefer2021transformer, hertz2022prompttopromptimageeditingcross}. Figure~\ref{fig:method} summarizes our analysis. We begin by asking whether specific proxy components - namely, particular Q-Former attention heads $h$
- exhibit a consistent preference for a single VLM layer, independent of its absolute index.

To investigate this, we define the total attention mass as:
$A_{h, q}^{(l)} = \sum_{i} a_{h, q}^{(l, i)}$, where $a_{h, q}^{(l, i)}$ is the attention score in Q-Former head $h$ from query $q$ to the $i$-th token in the $l$-th VLM layer.
We compute this aggregate mass across a diverse set of samples. Crucially, for each individual sample, we sort these layer-wise attention masses strictly by magnitude, effectively detaching the concentration of attention from the underlying VLM layer identity.

\input{figures/qformer-attn_icml}

Averaging these sorted magnitudes across all samples reveals a striking pattern (Figure~\ref{fig:layer-decay}). Specific proxy attention heads
act as ``decisive'' routers. On average, all queries within a decisive head assign significantly more attention mass to their top-ranked VLM layer than to the second-ranked one. This steep drop-off demonstrates a strong internal consensus for the informative VLM layer.

Because the dominant layer is input-dependent, we identify it on a per-sample basis. Let $\mathcal{H}$ denote the set of $H$ decisive heads. For any given input, we first extract the optimal layer $l^*_h$ for each individual head $h \in \mathcal{H}$ by finding the layer that receives the maximum total attention mass from its queries:
\vspace{-12pt}
$$l^*_h = \operatorname*{argmax}_{l} \sum_{q=1}^{N} A_{h, q}^{(l)}$$
\vspace{-4pt}

We then determine the overall optimal VLM layer for the input, $l^*$, by taking a majority vote across the preferred layers of all decisive heads:
$$l^* = \operatorname*{mode} \{l^*_h \mid h \in \mathcal{H}\}$$

Building on this layer-level routing, we scrutinize the token-level attention distribution exclusively within the identified dominant layer $l^*$.
Applying the same magnitude-based ranking to the attention scores $a_{h, q}^{(l^*, i)}$, for every query $q$ to VLM token $i$, in heads $h \in \mathcal{H}$, reveals an equally steep drop-off immediately following the highest-ranked token (Figure \ref{fig:token_decay}). As shown in Figure \ref{fig:token_significance}, these dominant VLM tokens are consistently semantically significant to the target edit.
Ultimately, this extreme sparsity leads to a crucial conclusion: the VLM's intermediate hidden states intrinsically encode highly concentrated, semantically grounded spatial representations, and the trained proxy essentially functions as a dynamic routing mechanism to retrieve them.

\subsection{Analysis of Spatial Information Flow in the VLM}
\label{subsec:vlm_analysis}

By leveraging the specific ``decisive'' attention heads of the Q-Former as a guide, we can effectively bypass the massive search space of the VLM's hidden states and directly pinpoint the representations that encode spatial information.

We analyze the internal attention maps of the dominant instruction tokens within the VLM, as identified by our proxy at the optimal layer $l^*$. For a given instruction token $i$ in the VLM, its attention scores to the image tokens $j$ are averaged across all heads in the VLM layer, yielding $\alpha_{i, j}^{(l^*)}$. A profound spatial correlation is evident when observing this metric: the instruction tokens most valued by the proxy directly and accurately attend to the visual subject of the edit.

This demonstrates a pronounced localization effect akin to the explicit grounding observed by Kaduri et al.~\cite{kaduri2024_vision_of_vlms} (see Figure \ref{fig:txt-to-image-attn}). Notably, our approach extracts this precise spatial localization entirely from the model's intermediate representations during a single forward pass. This circumvents the need for autoregressive token generation—a critical advantage, as the native editing pipeline inherently precludes the generation of new text tokens.

\input{figures/sem_dominance_and_attn_maps_icml}

%% file: figures/proxy_training_icml.tex
\begin{figure*}[t]
    \centering
    \begin{subfigure}[b]{0.48\linewidth}
        \centering
        \includegraphics[width=\linewidth]{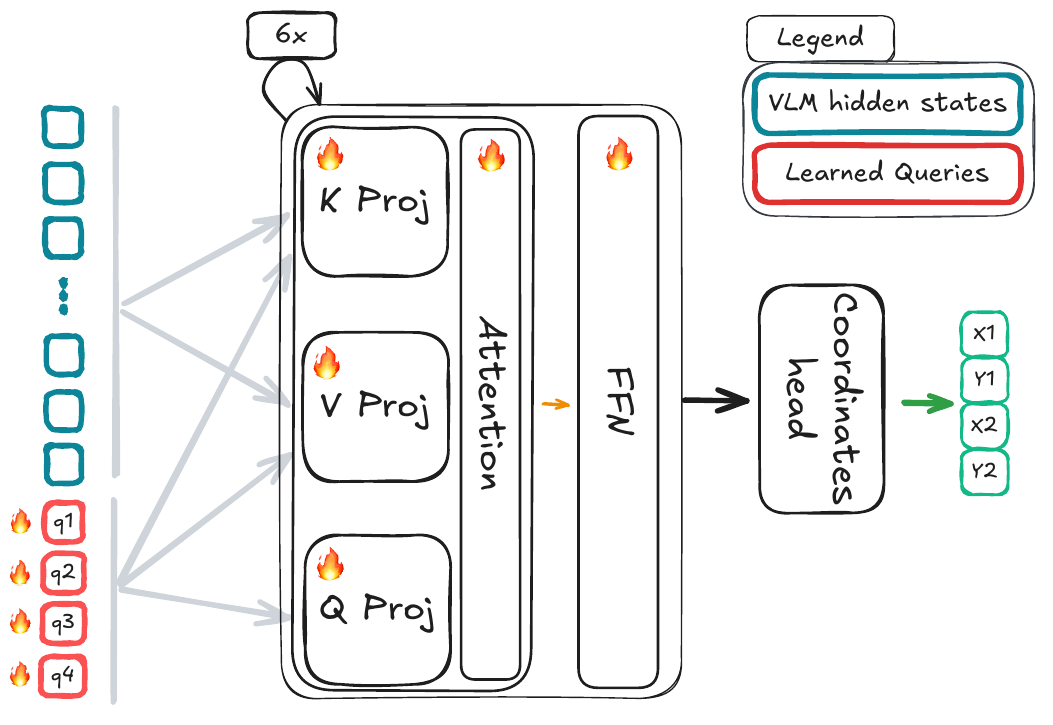}
        \caption{\textbf{Proxy model architecture.} VLM hidden states are projected exclusively to keys ($K$) and values ($V$). Only the learned queries are projected to $Q$, attending to all tokens across six transformer layers. Finally, a coordinate head maps the refined queries to bounding box coordinates $(x_1, y_1, x_2, y_2)$.}
        \label{fig:qformer_arch}
    \end{subfigure}
    \hfill %
    \begin{subfigure}[b]{0.48\linewidth}
        \centering
        \includegraphics[width=\linewidth]{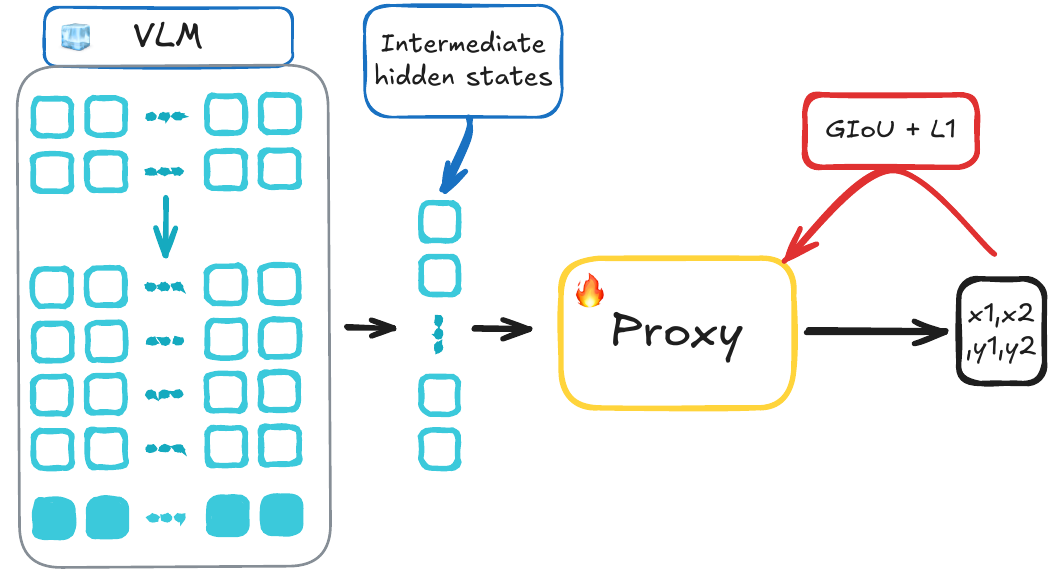}
        \caption{\textbf{Proxy training process.} The proxy receives VLM hidden states
        from a subset of layers
        $L$, $\{H^{(l)}\}_{l \in L}$.
        $\{H^{(l)}\}_{l \subseteq L}$.
        It predicts the spatial coordinates of the target edit region, optimized via $\mathcal{L}_1$ and $\mathcal{L}_{\text{GIoU}}$ losses. The ground truth bounding box is obtained by directly prompting the VLM and parsing its response.}
        \label{fig:proxy_train_fig}
    \end{subfigure}

    \caption{Proxy architecture (left) and the proxy training scheme (right).}
    \label{fig:proxy_training}
\end{figure*}
\vspace{-4pt}

%% file: figures/q-former_convergence_icml.tex
\begin{figure*}[t]
        \centering
        \includegraphics[width=0.6\linewidth]{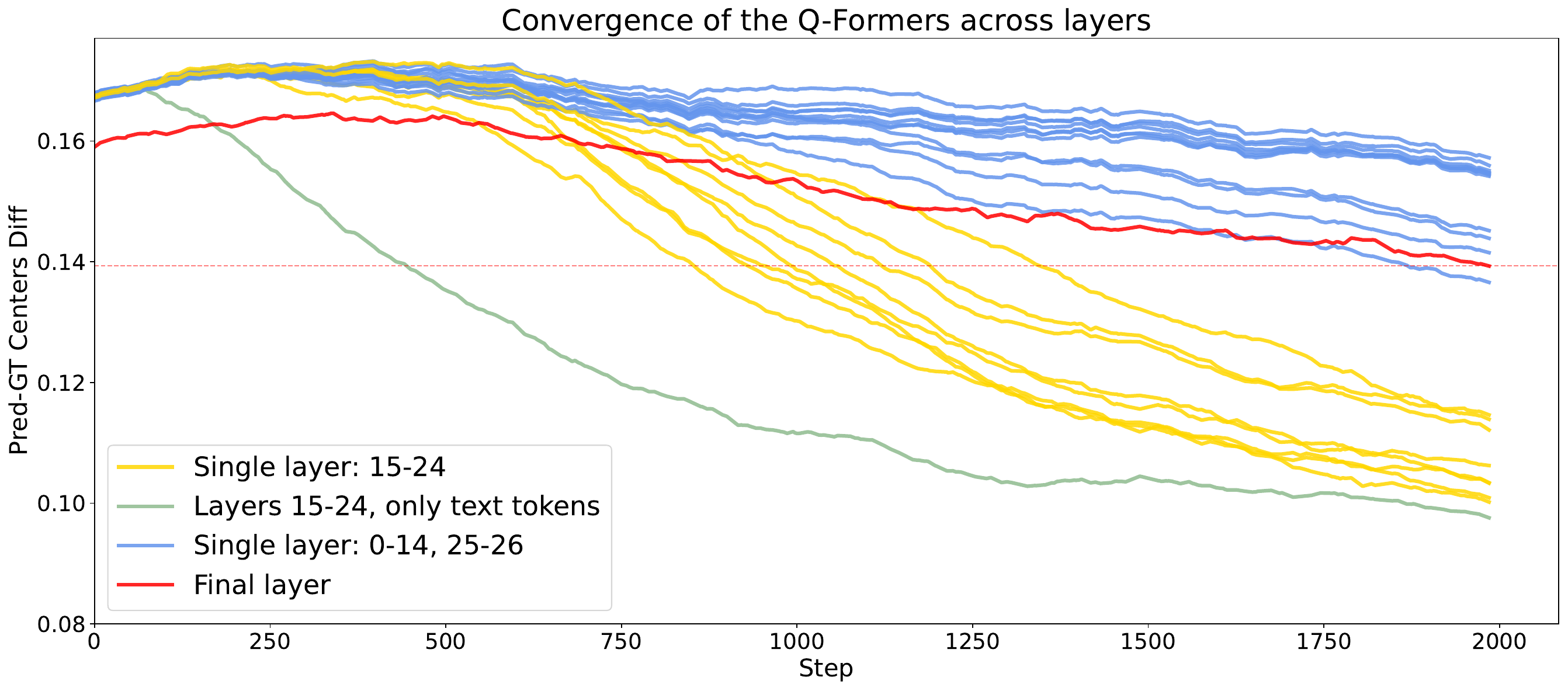}
        \caption{
        Mean localization error of the Q-Former proxy (measured by bounding box center distance) across different VLM layer configurations.
        The baseline model trained on the standard final-layer representations, $H^{(27)}$ (\textcolor{red}{\textbf{red}}), exhibits slow convergence. Similarly, proxies trained on early and very late layers, such as $l \in \{0\dots13, 25\dots26\}$ (\textcolor{blue}{\textbf{blue}}), demonstrate poor performance and struggle to converge. In contrast, training the proxy on the full sequence of hidden states from a single intermediate layer, $H^{(l)}$ (\textcolor{orange}{\textbf{yellow}}), yields notably faster convergence. Finally, restricting the proxy to attend exclusively to the specific hidden states of the user prompt tokens,
        $\{h_i^{(l)}\}_{l\in L}$, from middle layers
        $L=\{15,\dots,24\}$ (\textcolor{green!50!black}{\textbf{green}}),
        achieves the fastest convergence rate.
        } %
        \label{fig:qformer_converge}
\end{figure*}

%% file: figures/method_icml.tex
\begin{figure*}[t]
    \centering
    \includegraphics[width=0.93\linewidth]{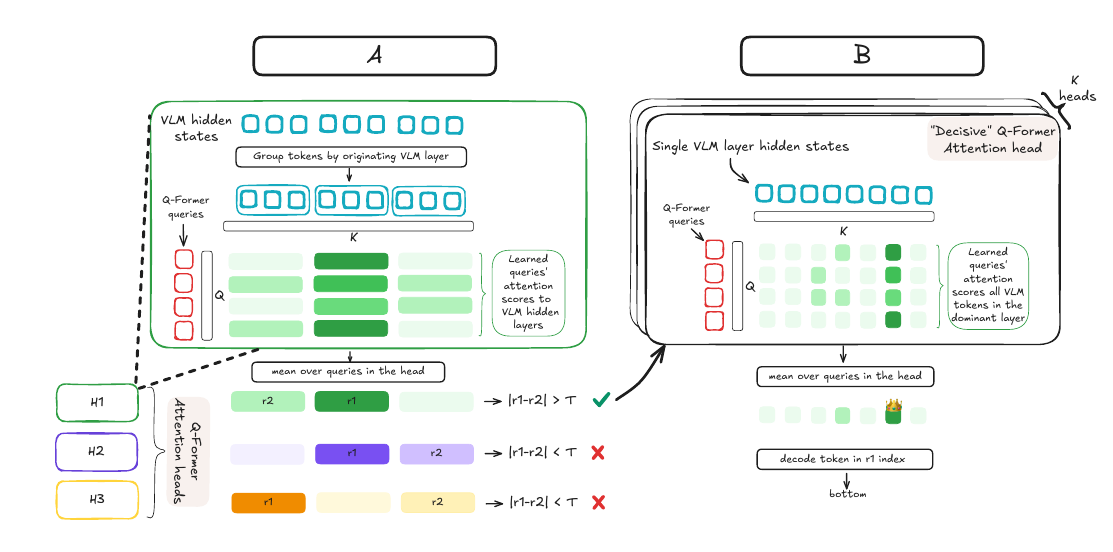}
    \caption{ \textbf{Decomposing the Proxy.} \textbf{(A)} We identify ``decisive'' attention heads in the Q-Former proxy, where each query assigns substantial weight to a single layer. While these layers are input-dependent, these heads maintain a sparse attention pattern ($|r_1 - r_2| > T$). \textbf{(B)} Within a decisive head and the dynamically identified VLM layer, we pinpoint the dominant hidden states anchoring the spatial signal. Decoding these reveals that they correspond to semantically significant nouns and locations that guide the edit (see Figure~\ref{fig:token_significance}).} %
    
    \label{fig:method}
\end{figure*}

%% file: figures/qformer-attn_icml.tex
\begin{figure*}[t]
    \centering
    \begin{subfigure}[b]{0.48\linewidth}
        \centering
        \includegraphics[width=\linewidth]{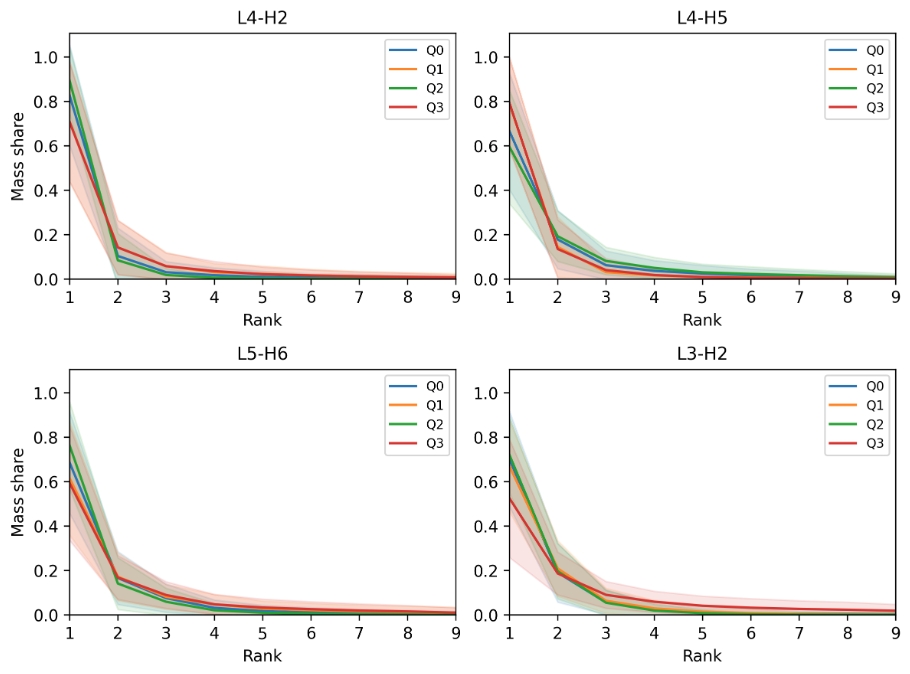}
        \caption{Layer-wise attention concentration. For specific Q-Former heads $h$,
        the queries $q$ concentrate their attention mass $A_{h, q}^{(l)}$
        heavily onto a single VLM layer $l$. Ranked layer-wise attention exhibits a steep decline, showing a mean drop of 0.6.
        This sparsity is consistent across all queries and samples.}
        \label{fig:layer-decay}
    \end{subfigure}
    \hfill %
    \begin{subfigure}[b]{0.48\linewidth}
        \centering
        \includegraphics[width=\linewidth]{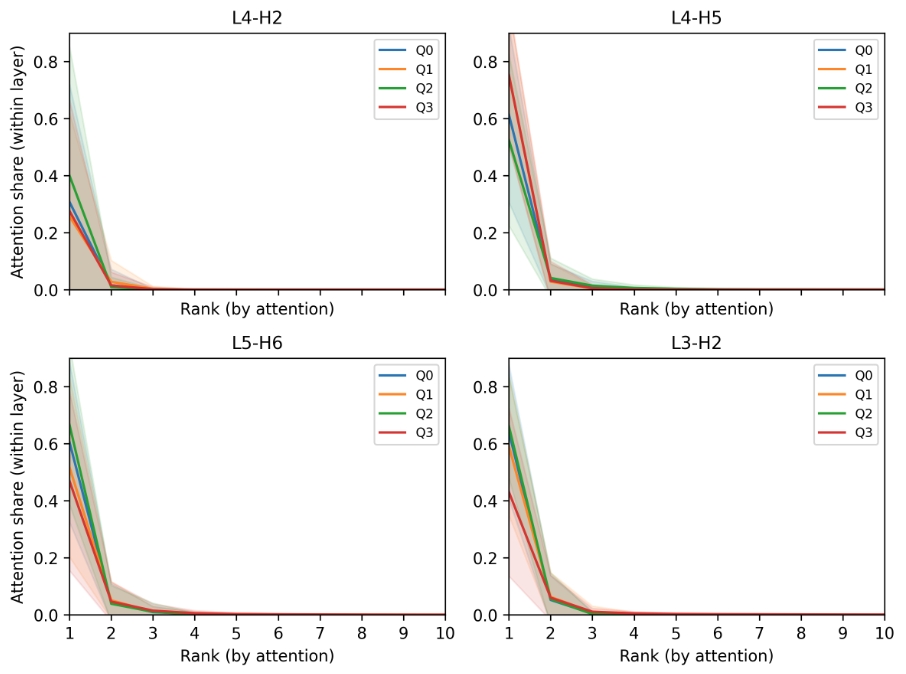}
        \caption{Token-wise attention concentration. Extending the analysis from to a token-level granularity reveals a similar sparsity. Within the identified dominant VLM layer $l^*$, these same Q-Former attention heads, $h$,
         concentrate their attention scores $a_{h, q}^{(l^*, i)}$
        onto a single target token index $i$ for each learned query $q$.}
        \label{fig:token_decay}
    \end{subfigure}

    \caption{Sparsity analysis of the Q-Former proxy's attention mechanisms, demonstrating highly localized attention patterns at both the layer and token levels.}
    \label{fig:qformer_attn}
\end{figure*}

%% file: figures/sem_dominance_and_attn_maps_icml.tex
\begin{figure*}[t]
    \centering
    \captionsetup[subfigure]{skip=1pt,font=footnotesize}
    \setlength{\abovecaptionskip}{3pt}
    \setlength{\belowcaptionskip}{0pt}

    \begin{subfigure}[t]{0.49\linewidth}
        \centering
        \setlength{\tabcolsep}{2pt}
        \renewcommand{\arraystretch}{0.9}
        \begin{tabular}[t]{@{}cc@{}}
            \includegraphics[width=0.44\linewidth]{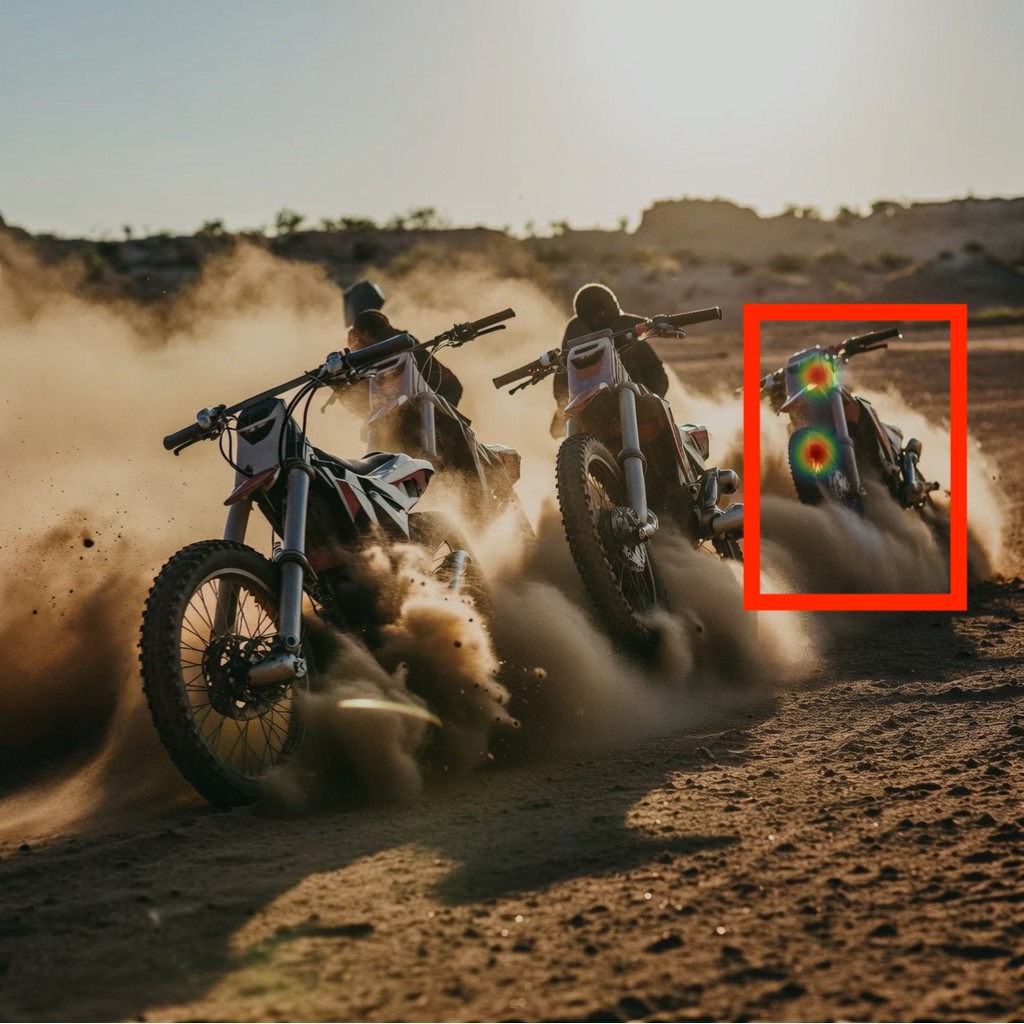} &
            \includegraphics[width=0.44\linewidth]{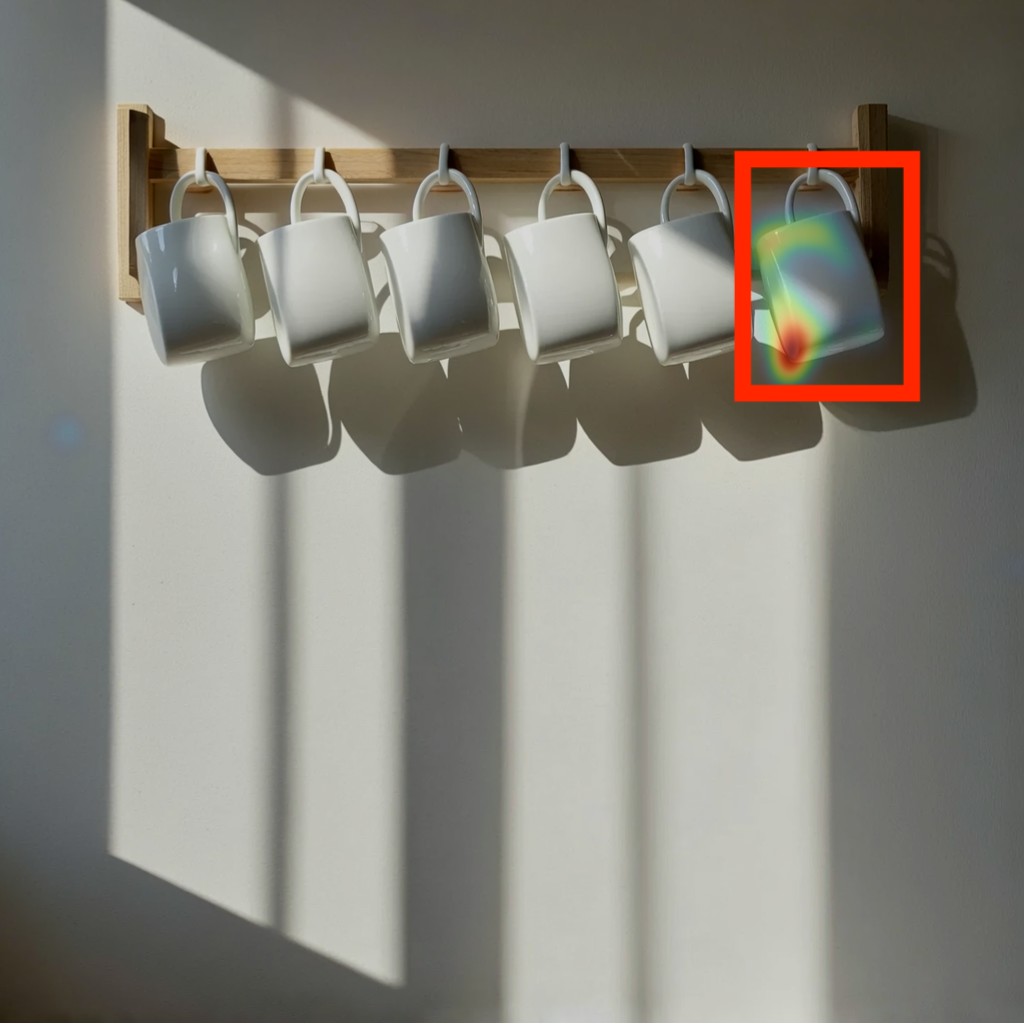} \\[-2pt]
            \parbox[t][0.95cm][t]{0.44\linewidth}{\centering \tiny ``Remove the dirt bike riding in the very back of the pack.''} &
            \parbox[t][0.95cm][t]{0.44\linewidth}{\centering \tiny ``Remove the white ceramic mug hanging on the far right of the scene.''} \\
            \includegraphics[width=0.44\linewidth]{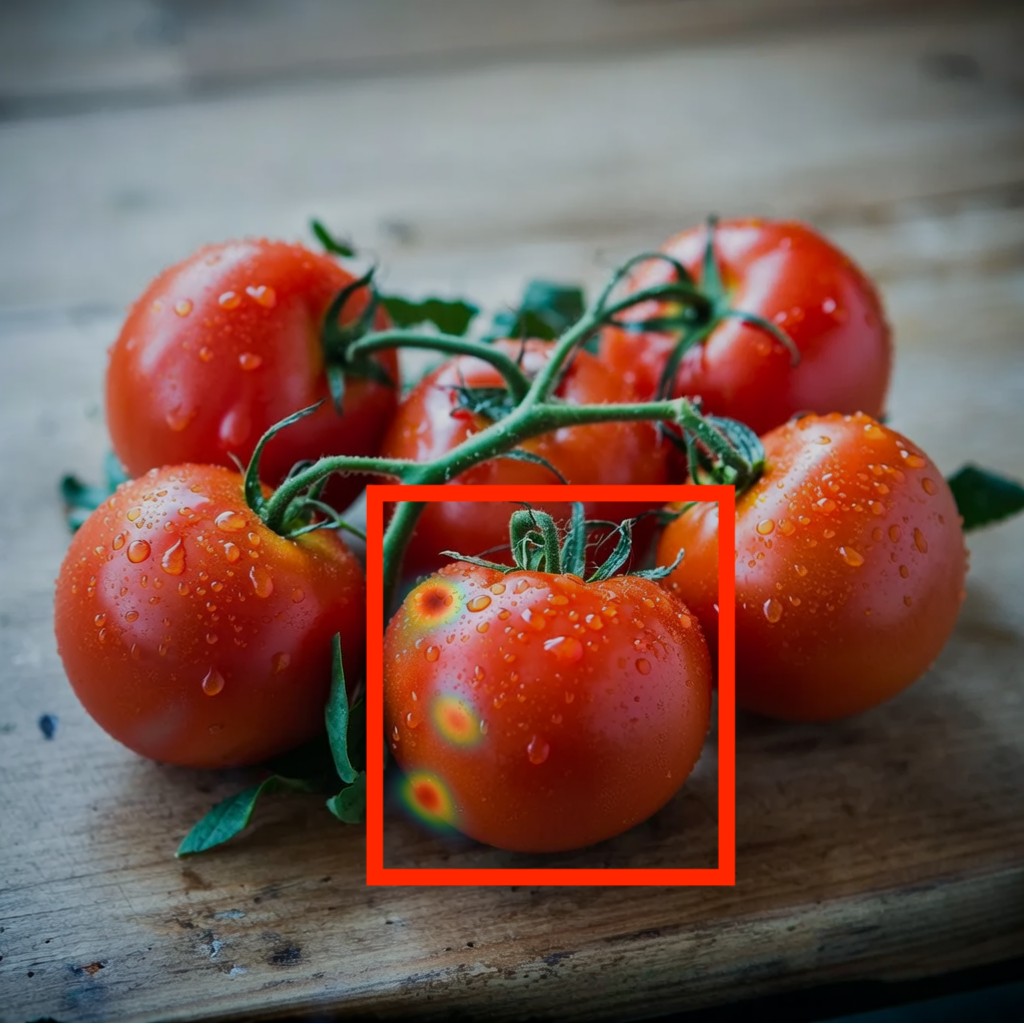} &
            \includegraphics[width=0.44\linewidth]{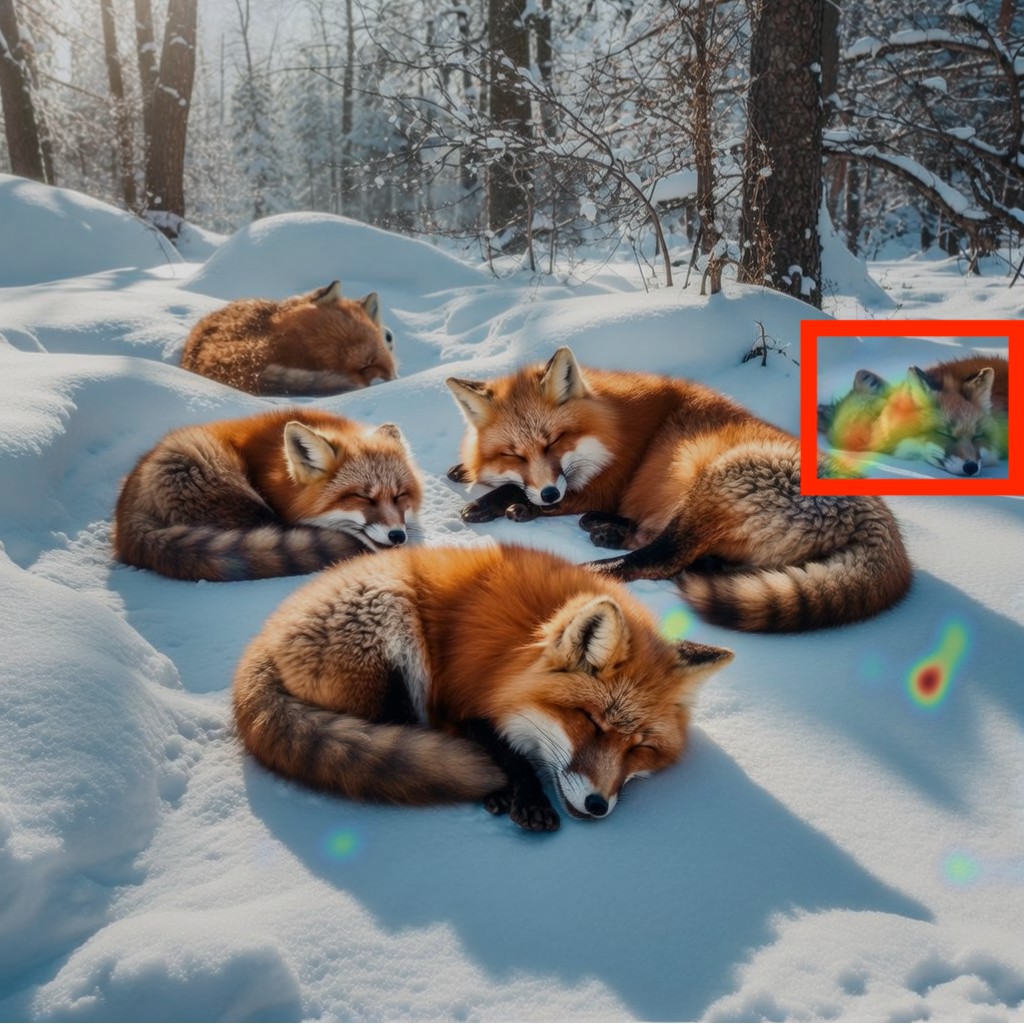} \\[-2pt]
            \parbox[t][0.95cm][t]{0.44\linewidth}{\centering \tiny ``Change the tomato closest to the front into a yellow heirloom tomato.''} &
            \parbox[t][0.95cm][t]{0.44\linewidth}{\centering \tiny ``Change the fur of the rightmost sleeping fox to a silvery-grey color.''} \\
        \end{tabular}
        \caption{Spatial attention maps of the dominant instruction tokens align with the edit targets.}
        \label{fig:txt-to-image-attn}
    \end{subfigure}
    \hspace{2pt}%
    \begin{subfigure}[t]{0.49\linewidth}
        \centering
        \setlength{\tabcolsep}{2pt}
        \renewcommand{\arraystretch}{0.9}
        \begin{tabular}[t]{@{}cc@{}}
            \includegraphics[width=0.44\linewidth]{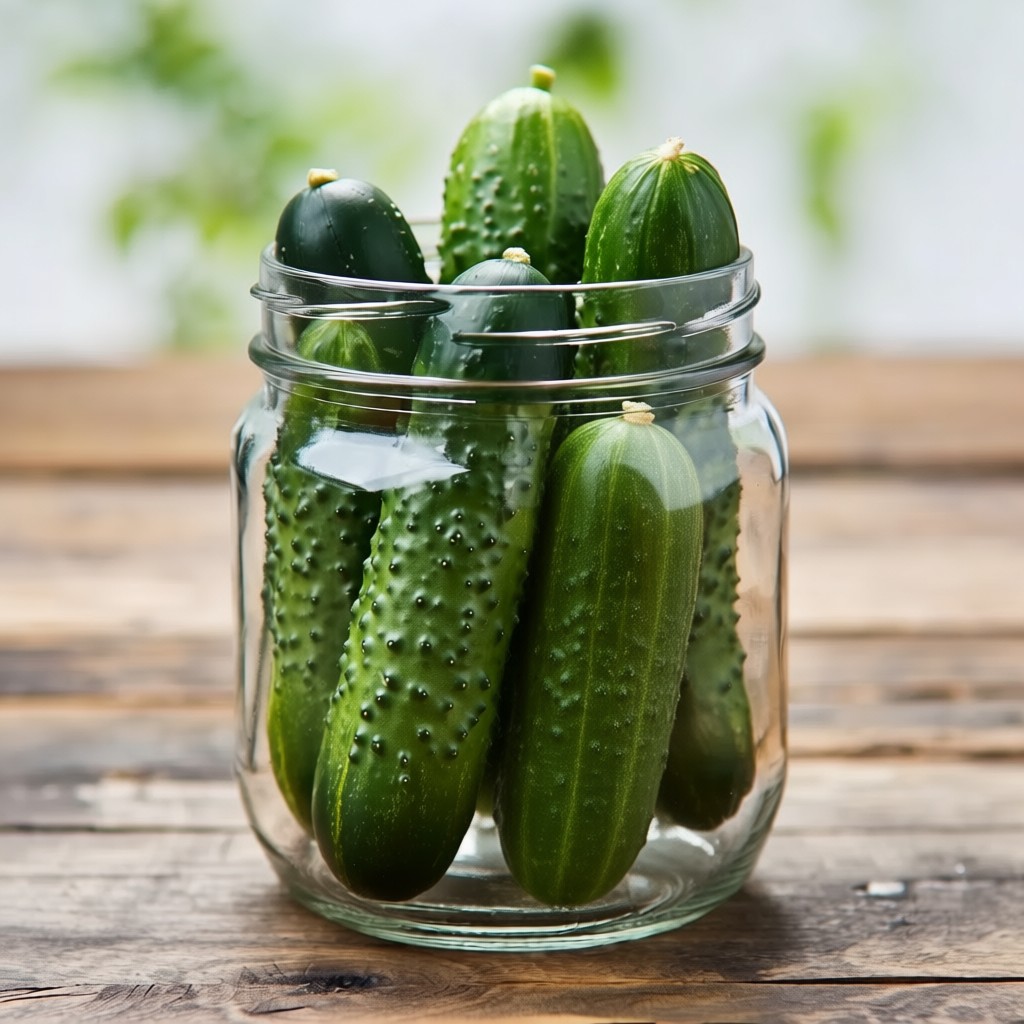} &
            \includegraphics[width=0.44\linewidth]{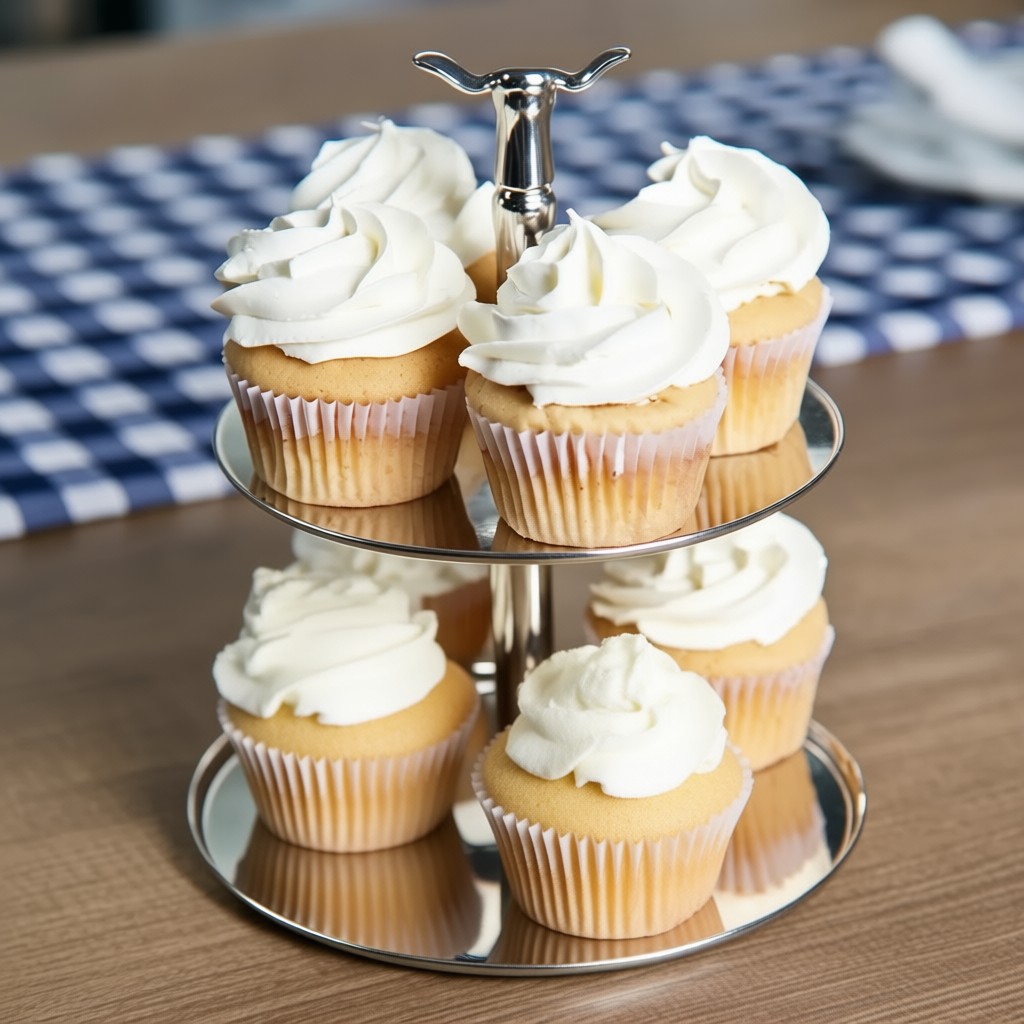} \\[-2pt]
            \parbox[t][0.95cm][t]{0.44\linewidth}{\centering \tiny ``Replace the front-right \textbf{cucumber (47\%)} with a pale yellow lemon cucumber.''} &
            \parbox[t][0.95cm][t]{0.44\linewidth}{\centering \tiny ``Replace the two front \textbf{cupcakes (53\%)} with red velvet cream cheese cupcakes.''} \\
            \includegraphics[width=0.44\linewidth]{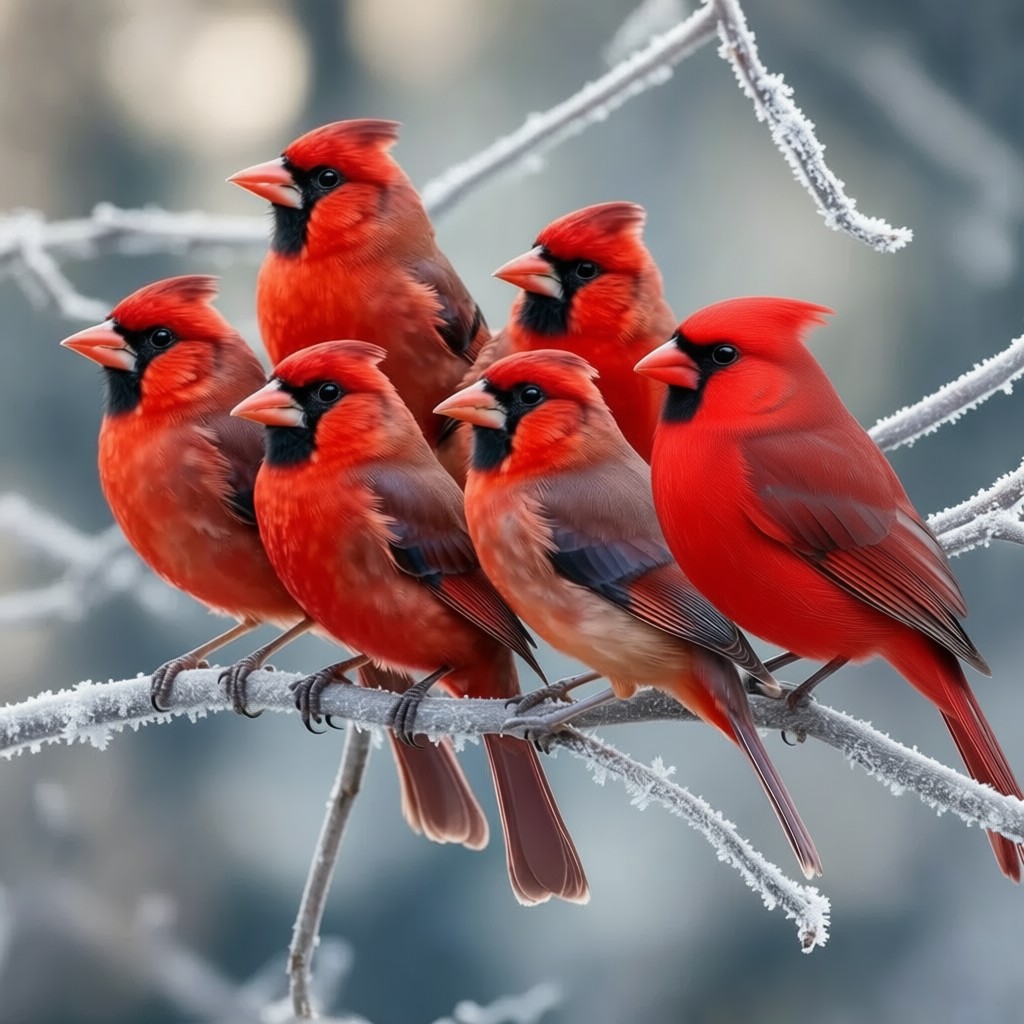} &
            \includegraphics[width=0.44\linewidth]{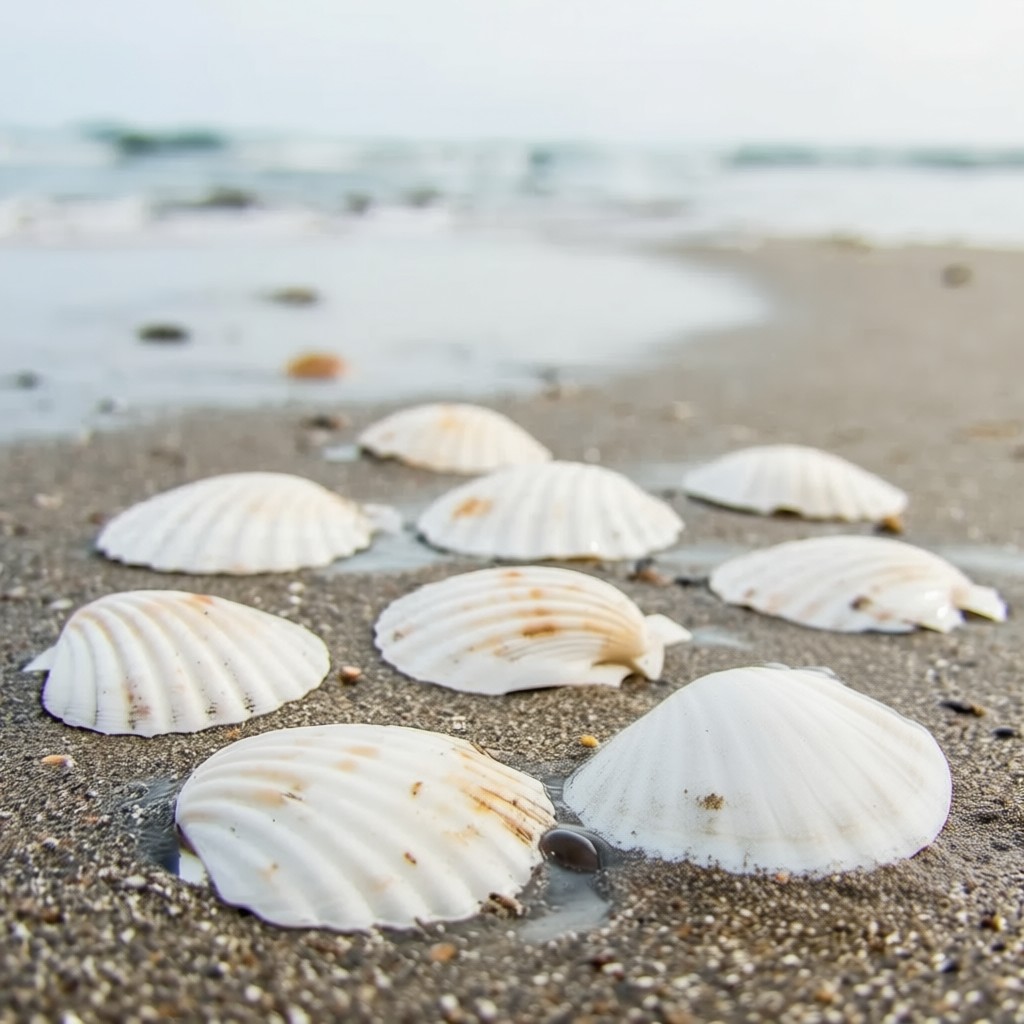} \\[-2pt]
            \parbox[t][0.95cm][t]{0.44\linewidth}{\centering \tiny ``Recolor the rightmost \textbf{bird (65\%)} to a completely solid shade of blue.''} &
            \parbox[t][0.95cm][t]{0.44\linewidth}{\centering \tiny ``Add a pink-lipped conch shell instead of the front \textbf{right (49\%)} scallop.''} \\
        \end{tabular}
        \caption{Within the dominant layer $l^*$, attention mass concentrates on the semantically significant tokens (attention share in bold).}
        \label{fig:token_significance}
    \end{subfigure}

    \caption{Analysis of VLM internal representations across multiple editing instructions.} %
    \label{fig:vlm_analysis_combined}
\end{figure*}

%% file: sections/5_clean_improving_icml.tex
\subsection{Validating the Recovered Localization Signal in the Editing Pipeline}

Our analysis reveals an untapped potential for improving localization by recovering spatial signals that, while degraded in late VLM layers, remain preserved in earlier representations.
To leverage these signals during inference,
we employ the proxy to generate a bounding box for the target object described in the edit prompt. Examples of such predicted boxes alongside the corresponding failures of the original Qwen-Image-Edit model are provided in  Figure \ref{fig:localization}.

To condition the DiT on the proxy's spatial output, we fine-tune the model via a LoRA \cite{hu2021loralowrankadaptationlarge} module to recognize overlaid bounding boxes as explicit localization cues (see Supplementary Material for dataset curation details). During inference, our pipeline operates in two stages. First, the single VLM forward pass extracts intermediate hidden states, which the Q-Former uses to predict the target's bounding box. 
This predicted box is then visually overlaid onto the source image and encoded into the DiT's latent space via the VAE. 
Guided by both the textual instruction and this newly introduced spatial marker, the DiT accurately localizes the modification while learning to edit out the box artifact from the final generated output.

\input{figures/qualitative_icml}

\vspace{-2pt}
\paragraph{\textbf{Evaluation}}
We evaluate our approach with Qwen-Image-Edit 
against several established image editing pipelines and alternative conditioning strategies. Specifically, we compare our performance to state-of-the-art models including FLUX-Kontext \cite{labs2025flux1kontextflowmatching}, FLUX.2 \cite{flux-2-2025}, and FIBO-Edit \cite{gutflaish2025generatingimage1000words}, and the baseline Qwen-Image-Edit \cite{wu2025qwen}. %

To assess the efficacy of our conditioning method, which extracts spatial bounding boxes from intermediate VLM hidden states, we implement two additional conditioning variants within the baseline Qwen-Image-Edit pipeline.
The first variant utilizes a full autoregressive scene description. In this setup, the standard system prompt directs the VLM to exhaustively detail the image and its constituent elements. The resulting comprehensive text is generated autoregressively and is subsequently re-encoded in a second forward pass to serve as the conditioning signal.
The second variant implements the norm-averaging technique proposed by Wang et al. \cite{wang2025comprehensivestudydecoderonlyllms}. This method aggregates internal representations across layers to form the condition.

\paragraph{\textbf{Quantitative Results}}
We evaluate semantic editing success using a Vision Question Answering (VQA) approach~\cite{lin2024evaluatingtexttovisualgenerationimagetotext} via Gemini 2.5 Pro~\cite{gemini25report2025}. Unlike global metrics (e.g., CLIP) that often miss localized nuances, VQA allows to explicitly assess targeted semantic changes such as subject modification and instruction adherence 
As a structural complement, we measure LPIPS~\cite{zhang2018unreasonableeffectivenessdeepfeatures} exclusively outside the target's ground-truth bounding box. This enables to quantify background preservation and verify that the edit introduces no unintended artifacts to the surrounding scene. As shown in Figure~\ref{fig:quantitative}, our method achieves the highest mean VQA score while maintaining low background LPIPS, indicating semantically accurate and highly localized edits. Notably, the autoregressive ``Full Description'' variant is the strongest baseline, corroborating our hypothesis that sequential decoding helps to surface important spatial cues that remain dormant during a single forward pass.
\input{figures/quantitative_icml}

\paragraph{\textbf{Qualitative Results}}
Qualitative results are demonstrated in Figure~\ref{fig:qual_results}. Providing this explicit spatial condition resolves a variety of localization failures. While the standard pipeline often struggles to alter the correct object, our guided approach correctly focuses the DiT, ensuring that visual changes are applied strictly to the intended target. This effectively reconciles the discrepancy between the VLM's internal knowledge and the final generated output.

%% file: figures/qualitative_icml.tex
\begin{figure*}[t]
    \centering
    \begin{minipage}{0.95\linewidth}
    \centering
    \scriptsize
    \setlength{\tabcolsep}{1.5pt} %

    \begin{tabular}{c c c c c c c c}
        \tiny \textbf{Input} & \tiny \textbf{Ours} & \tiny \textbf{Baseline} & \tiny \textbf{Kontext} & \tiny \textbf{FIBO} & \tiny \textbf{FLUX.2} & \tiny \textbf{Full desc} & \tiny \textbf{Norm-AVG} \\[6pt]

         \multicolumn{8}{c}{\tiny ``Change the hue of the lowest quartz cluster to a deep amethyst purple.''} \\[2pt]
        \includegraphics[width=0.115\linewidth]{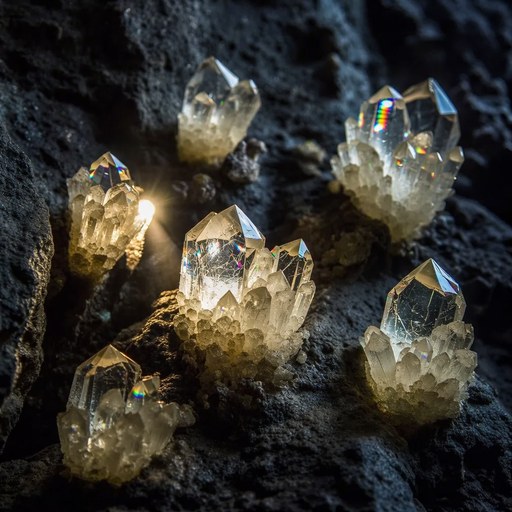} &
        \includegraphics[width=0.115\linewidth]{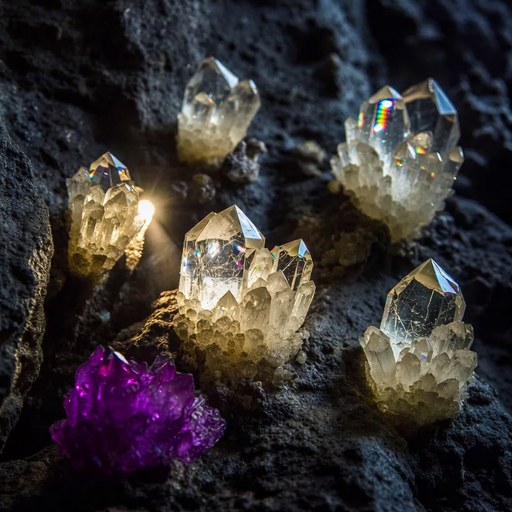} &
        \includegraphics[width=0.115\linewidth]{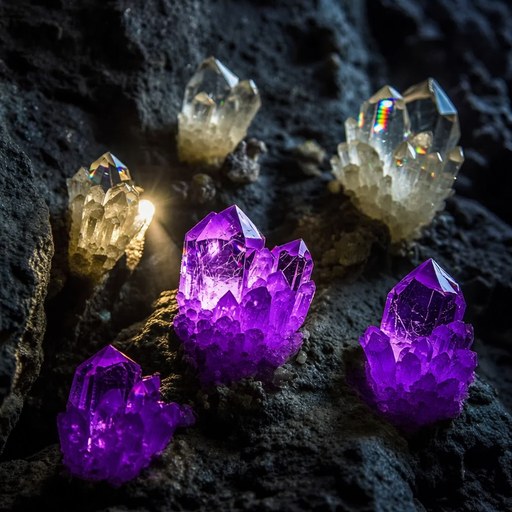} &
        \includegraphics[width=0.115\linewidth]{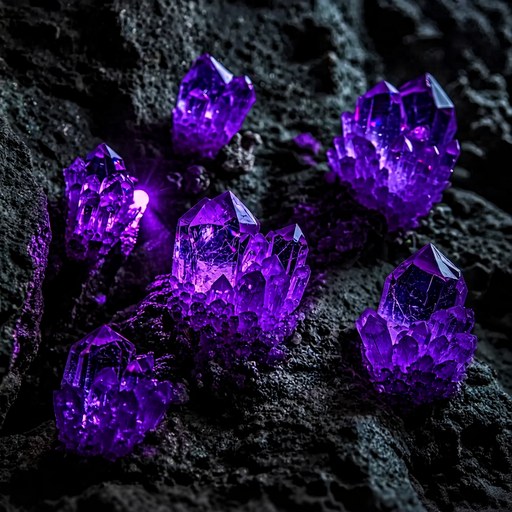} &
        \includegraphics[width=0.115\linewidth]{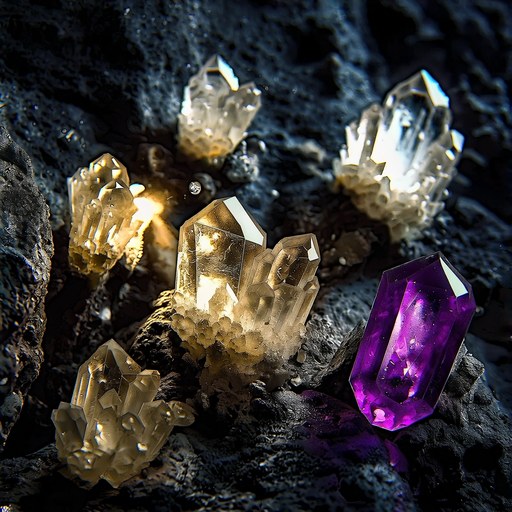} &
        \includegraphics[width=0.115\linewidth]{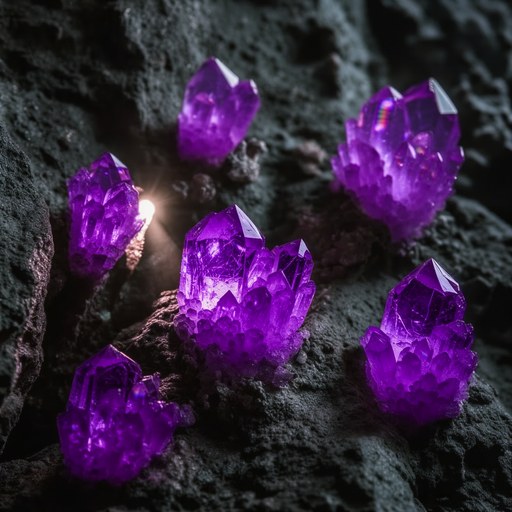} &
        \includegraphics[width=0.115\linewidth]{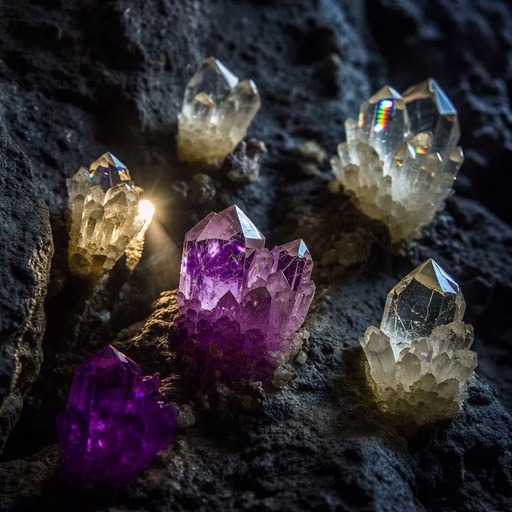} &
        \includegraphics[width=0.115\linewidth]{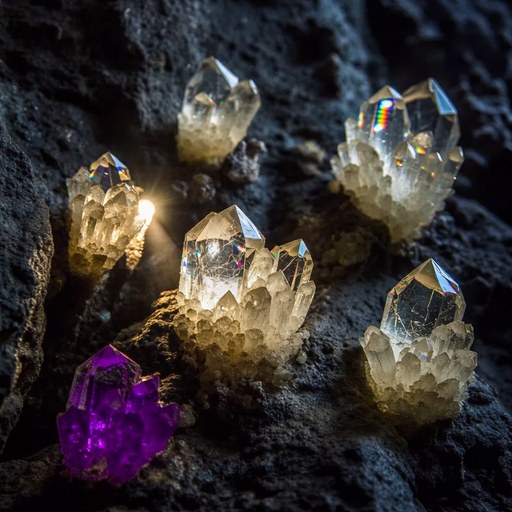} \\[3pt]

        \multicolumn{8}{c}{\tiny ``Add a slice of floating yellow lemon to the water in the leftmost jar.''} \\[2pt]
        \includegraphics[width=0.115\linewidth]{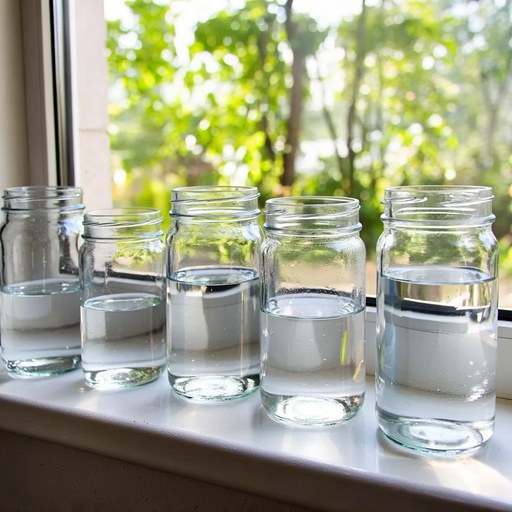} &
        \includegraphics[width=0.115\linewidth]{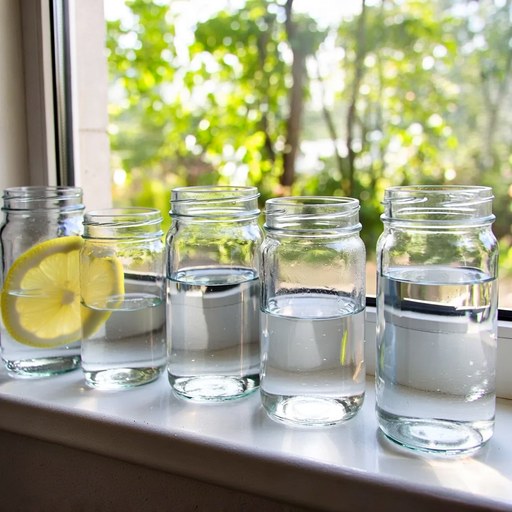} &
        \includegraphics[width=0.115\linewidth]{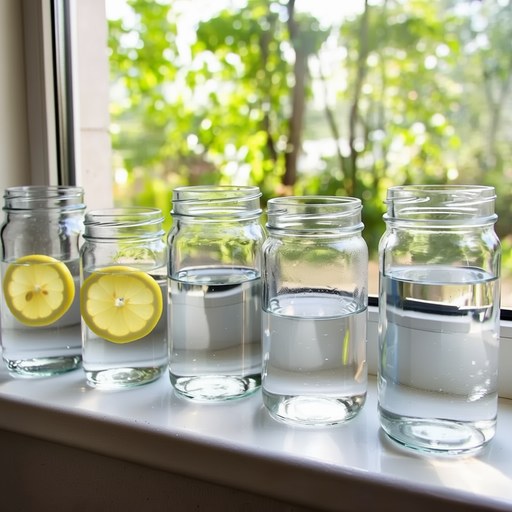} &
        \includegraphics[width=0.115\linewidth]{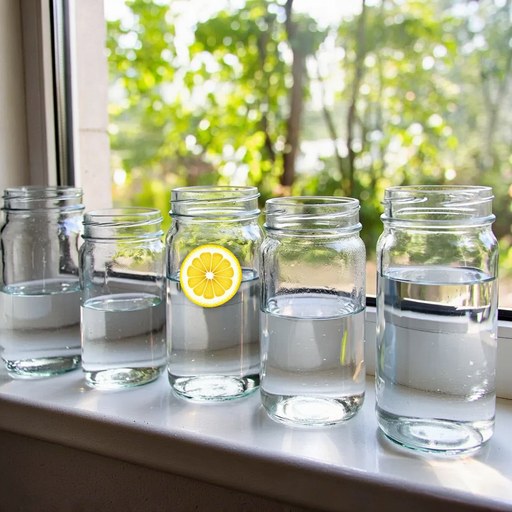} &
        \includegraphics[width=0.115\linewidth]{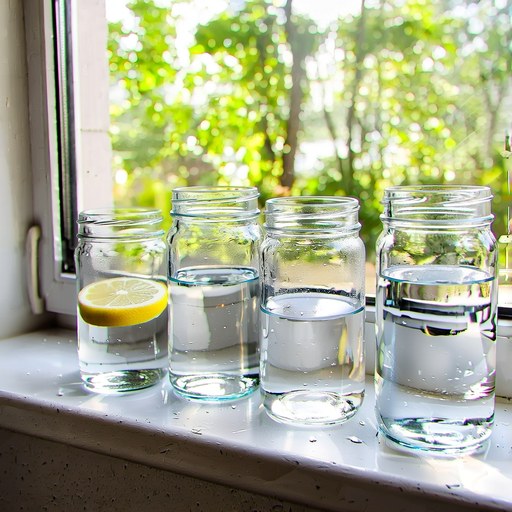} &
        \includegraphics[width=0.115\linewidth]{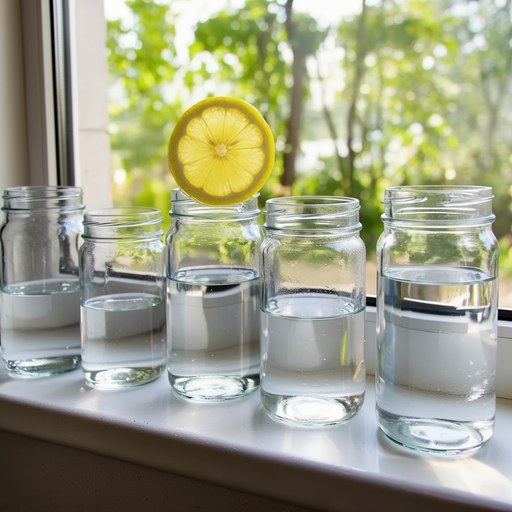} &
        \includegraphics[width=0.115\linewidth]{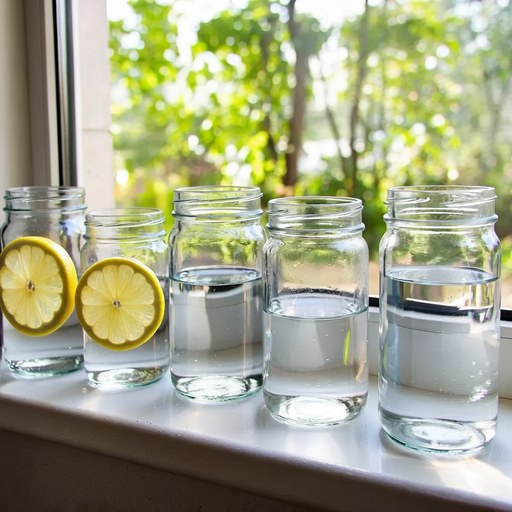} &
        \includegraphics[width=0.115\linewidth]{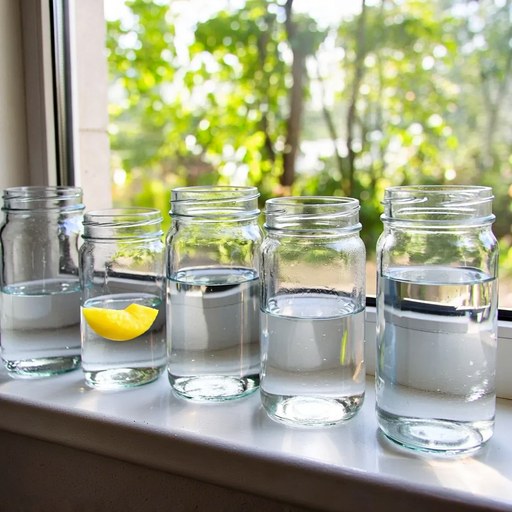}

    \end{tabular}
    \end{minipage}

    \caption{Qualitative comparison of image editing results. While existing methods frequently modify incorrect objects or apply changes broadly across the scene, our method successfully localizes the edit to the intended target. This demonstrates that providing explicit spatial conditioning enables more precise and reliable edit localization. Additional qualitative comparisons are provided in Figure \ref{fig:qual_results_supp}.} %
    \label{fig:qual_results}
\end{figure*}
\vspace{-5pt}

%% file: figures/quantitative_icml.tex
\begin{figure}[htbp]
    \centering
    \includegraphics[width=\linewidth]{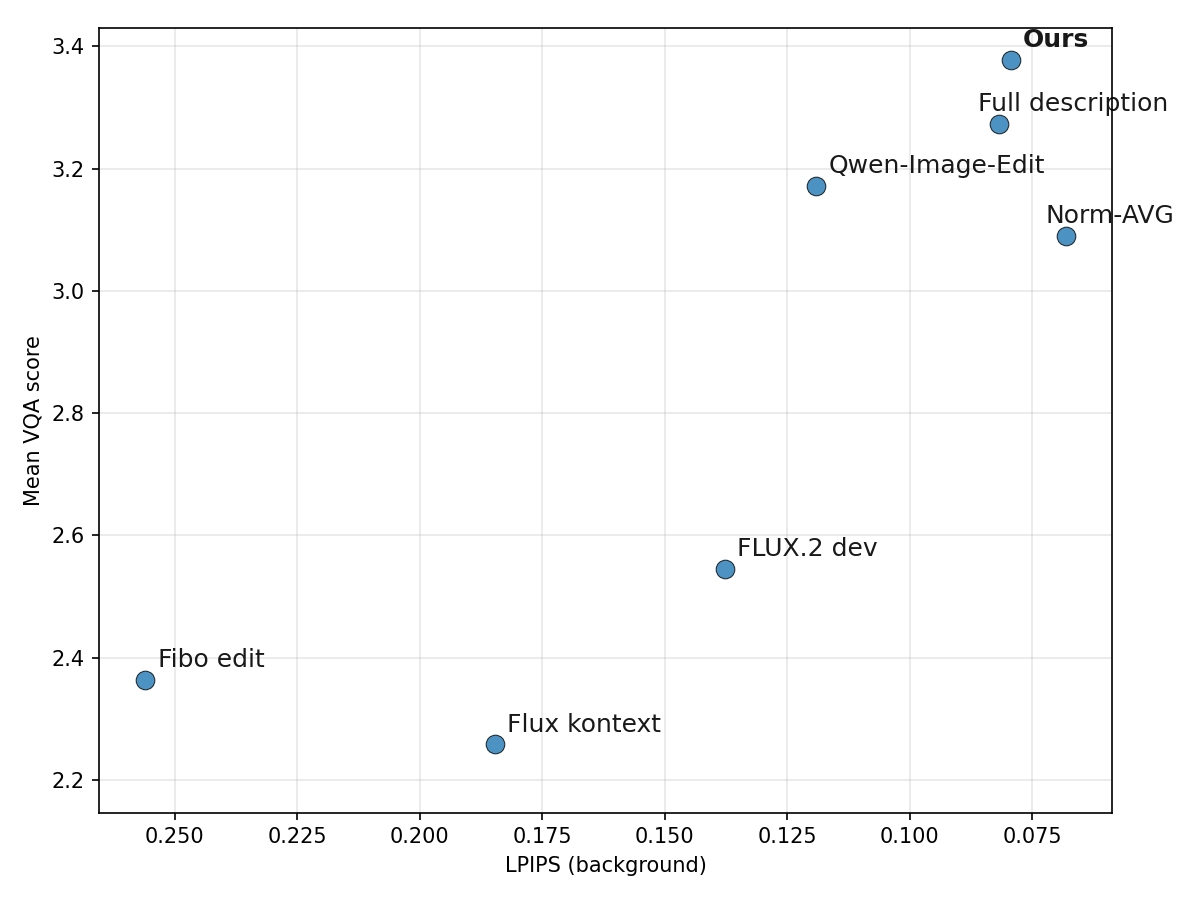}
    \caption{The horizontal axis shows LPIPS measured outside the ground-truth bounding box (LPIPS background), capturing structural changes to the surrounding scene. The vertical axis shows the mean VQA score (Gemini 2.5 Pro) evaluating subject modification, background cleanliness, and overall instruction adherence. Our method achieves the highest mean VQA score among all methods while maintaining low background distortion as measured by LPIPS.}
    \label{fig:quantitative} %
\end{figure}

%% file: sections/7_conclusion_icml.tex
\section{Conclusions}

In this work, we analyze Vision-Language Models (VLMs) serving as single-pass condition encoders.
We propose a lightweight framework for analyzing VLM mechanisms in a non-generative setting.
Our analysis reveals that current image editing pipelines under-utilize the spatial information encoded within these models; specifically, we demonstrate that rich, highly precise localization signals peak in intermediate representations and are harder to decode from the final layer. 
\newline
\noindent{We empirically validate these findings through a minimal modification that successfully recovers and integrates these intermediate signals to correct downstream editing failures.}

More broadly, our work highlights a critical design choice in multimodal pipelines: the specific operational mode used to extract information from the conditioning model.
We argue for a different approach to navigating this conditioning design space. Our strategy involves first identifying where task-relevant information resides within the model's internal layers and then distilling it into a compact, specialized signal. This targeted approach enables precise downstream performance while maintaining architectural efficiency.

This work opens several research avenues for better integration of VLMs into the generative pipeline.
First, our analysis framework can be extended to other tasks and architectures to further characterize internal model mechanisms. Additionally, it can be used to develop principled methods that better balance the rich information of autoregressive inference with the efficiency of a single forward pass.
Second, future research could explore training generative models with a more granular conditioning philosophy, where fixed layer-level hidden states are no longer the primary unit of conditioning. Under this paradigm, the network could learn an input-dependent conditioning structure, such as dynamically selecting representations across layers or pruning tokens based on the specific input.

\paragraph{Limitations.} 
Our proposed editing solution is intentionally straightforward. While the gains are consistent, there is still room to explore more expressive conditioning mechanisms. Currently, our implementation is limited to a single editing pipeline, as it is the only one that aligns with our focus—a VLM that jointly conditions on text and images. However, the framework is designed to be easily adaptable to future models for automated interpretability.

%% file: sections/8_supp_icml.tex
In this supplementary material, we provide additional details on the datasets (\S\ref{supp:data-sec}), implementation (\S\ref{supp:implementation-details}), and further experimental results (\S\ref{supp:results}).

\section{Datasets Details}
\label{supp:data-sec}

\subsection{Ground Truth Bounding Box Extraction}
\label{lab:gt-bbox}

The experiment in Section 4.1 of the main paper and the data generation pipeline used for model fine-tuning in Section 5 require accurate bounding boxes for an input image and an edit prompt.
To obtain these ground truth bounding boxes, we use Qwen2.5-VL-7B-Instruct.
We choose this model as it also serves as the text backbone in Qwen-Image-Edit, ensuring consistency between the VLM used for editing and for bounding box prediction.
We load the model in \texttt{bfloat16} precision and provide it with the source image along with the following system prompt:

\begin{tcolorbox}[colback=gray!10, colframe=gray!50, boxrule=0.5pt]
\ttfamily
Given the image and the user's text instruction, identify the object or objects that are the subject of this edit instruction, then explain how the user's text instruction should alter or modify the image. Generate a bounding box that meets the user's requirements for the edit. Do not include any other text or formatting, where (x1, y1) is the top-left corner and (x2, y2) is the bottom-right corner.
\end{tcolorbox}
We run the generation deterministically without sampling and cap it at 400 new tokens to ensure consistent outputs.

Once the model generates the text, we parse it to extract the four bounding box coordinates: $x_1$, $y_1$, $x_2$, and $y_2$. To ensure the coordinates form a valid box, we enforce $x_1 < x_2$ and $y_1 < y_2$ by swapping values when necessary. All coordinates are clamped to the image width and height to prevent out-of-bounds bounding boxes.

\subsection{Training and Evaluation Dataset Generation}
\label{supp:data}

Most existing open-source editing datasets primarily focus on single-object images. Although a few datasets address multi-object scenes, we found them to be overly noisy for the requirements of our task. Specifically, these datasets contain very few samples requiring spatial referencing among semantically similar objects—a necessary condition for isolating the phenomenon we aim to investigate. To address the lack of a suitable dataset, we construct our own dataset. We use this dataset for both training and evaluation.

To construct the dataset, we employ a multi-step pipeline to create paired source and target images together with their corresponding edit instructions.

First, we use Gemini 2.5 Pro \cite{gemini25report2025} to generate a set of prompt pairs. Each pair consists of (1) a prompt describing a scene containing a single anomalous object among otherwise uniform items (e.g., three dogs in a row where one is a different breed), and (2) an edit prompt describing a transformation that changes the anomalous object so that it matches the rest of the scene (e.g., turn the different dog into the same breed as the others). We refer to the first prompt as the \textit{generation prompt} and the second as the \textit{edit prompt}. The system prompt used to generate this dataset is provided in Appendix \ref{app:dataset_prompts}.

Second, we generate the paired images. We synthesize the first set of images from the generation prompts using FIBO \cite{gutflaish2025generatingimage1000words} with its standard runtime configuration. We choose FIBO due to its strong adherence to spatial layouts.
To generate the corresponding paired images, we use Qwen-Image-Edit-2509, guided by the edit prompts, which instruct the model to modify the anomalous object so that it matches the rest of the scene. This editing step, which harmonizes a single outlier with its surroundings, benefits from strong contextual guidance, as the neighboring uniform objects provide a clear semantic reference.

This process yields high-quality image pairs. For the actual training task, we reverse this relationship: the uniform edited image serves as the input (source), and the original anomalous image serves as the desired output (target). To obtain the text instruction for this reversed transformation, we use Mistral-Small-3.2-24B-Instruct-2506. We concatenate the target and source images side-by-side and provide them to the model with the following system prompt:

\begin{tcolorbox}[colback=gray!10, colframe=gray!50, boxrule=0.5pt]
\ttfamily
You are an expert at describing image edits. You will see two images side by side: LEFT = TARGET (desired result), RIGHT = SOURCE (the image to be modified). TASK: Write one edit instruction that transforms the RIGHT image into the LEFT image. \dots Use ONLY [spatial terms]: leftmost, rightmost, second from the left, \dots frontmost, backmost, \dots Output format: Turn the [EXACT\_SPATIAL\_POSITION] [object] into [detailed description]. NEVER identify the object by appearance/color/breed/type/size, ONLY by spatial position.
\end{tcolorbox}

We generate these edit prompts with a maximum of 256 new tokens, a temperature of 0.1, and a repetition penalty of 1.1. This process yields specific, spatially grounded instructions that map the uniform source images back to the anomalous target images. These image pairs and text instructions, together with ground-truth bounding boxes extracted using the process described in the Section \ref{lab:gt-bbox}, constitute the final training dataset.

For both train and evaluation datasets, we also generate ground-truth bounding boxes, as detailed in \ref{lab:gt-bbox}. The boxes are manually verified by a human annotator. All prompts and descriptions are also verified by a human annotator. We split the dataset 50\%-50\% between train and evaluation.

\section{Implementation Details}
\label{supp:implementation-details}

\subsection{Q-Former Architecture and Training Procedure}

We now provide details on the Q-Former trained in Section 4 of the main paper, which serves as our bounding box predictor. The input to the Q-Former is designed to simulate the instruction input to the DiT in the editing pipeline, namely the output of the VLM given an input image and an instruction prompt.
Therefore, to train the Q-Former we need a dataset containing images along with edit instruction that are suitable for them, and a bounding box for the object that should be edited.

To train the Q-Former, we require a dataset containing images, corresponding edit instructions, and bounding boxes indicating the object to be edited.
To ensure the Q-Former remains as general as possible, we leverage large-scale training data rather than our own specialized dataset. Specifically, we utilize the training split of the TIGER-Lab/OmniEdit-Filtered-1.2M dataset~\cite{wei2025omnieditbuildingimageediting}. This dataset consists of 1.2 million high-resolution image editing pairs across seven distinct tasks, including object swapping, removal, and style transfer. We use $128,000$ samples from this dataset.
For the ground-truth bounding boxes, we rely on the procedure described in Section~\ref{lab:gt-bbox}.
To ensure data quality, we filter the training samples to exclude bounding boxes with extreme normalized dimensions, restricting the area to a minimum of 0.001 and a maximum of 0.9.

During training, we feed the source image and edit prompt into Qwen2.5-VL-Instruct to extract hidden states from the target layers, following the same procedure used in the editing pipeline. Consistent with the editing pipeline, we filter out the system prompt tokens and retain the remaining sequence.

For the training objective, we use a combination of $L_1$ loss and Generalized Intersection over Union (GIoU) \cite{Rezatofighi_2018_CVPR} loss. To stabilize the initial training phase, we apply a warmup period of 100 steps during which the model is optimized using only the $L_1$ loss. After this warmup, we introduce the GIoU loss with a fixed weighting factor $\lambda = 0.4$, resulting in the following combined loss function:

$$ \mathcal{L}_{\text{total}} = (1 - \lambda) \mathcal{L}_{L_1} + \lambda \mathcal{L}_{\text{GIoU}} $$

The model is trained for a single epoch using the Adam optimizer with a base learning rate of 1e-4. The learning rate follows a cosine decay schedule, dropping to 30\% of its base value after the warmup period. We use a batch size of 64 and process 2000 batches.

\subsection{LoRA Fine-Tuning Details}
\label{supp:lora}

We provide details on the LoRA module fine-tuned on top of Qwen-Image-Edit~\cite{wu2025qwen} to consume the bounding-box overlay produced by the Q-Former at inference time.

\paragraph{Training data.}
The LoRA is trained on the curated triplet dataset described in Section~\ref{supp:data} (source image, edit instruction, edited image), paired with the human-verified ground-truth bounding boxes obtained as described in Section~\ref{lab:gt-bbox}. The LoRA is fine-tuned on this smaller, spatially curated set in order to teach the DiT to consume the overlaid box as a strict localization cue. 

\paragraph{Bounding-box overlay preprocessing.}
During training, we augment each source image by rendering its ground-truth bounding box directly onto the RGB pixels prior to encoding the image into the DiT's latent space. The target image (the edited image) is left unmodified, so the network must learn to both attend to the box as a spatial cue and remove the artifact from its prediction.

\paragraph{LoRA configuration.}
We attach LoRA adapters to the attention projection layers of the DiT: $q$, $k$, $v$, and the output projection (\texttt{to\_q}, \texttt{to\_k}, \texttt{to\_v}, \texttt{to\_out.0}). We use rank $r=16$ and scaling factor $\alpha=16$, with Gaussian initialization for the adapter weights. All other weights of Qwen-Image-Edit (VAE, base DiT, text encoder) remain frozen.

\paragraph{Optimization.}
The LoRA is trained with AdamW ($\beta_1=0.9$, $\beta_2=0.999$, $\varepsilon=10^{-8}$, weight decay $0.01$) at a base learning rate of $1{\times}10^{-4}$ under a cosine schedule with $50$ warmup steps. We use a per-device batch size of $2$ , and train for up to $3000$ steps over $15$ passes through the data. The training objective is the standard flow-matching loss of Qwen-Image-Edit. We train in \texttt{bfloat16} mixed precision with gradient checkpointing enabled and clip gradient norms at $1.0$. We take the best checkpoint per method.

\paragraph{Inference pipeline.}
\label{supp:editing-inference}
At inference, given a source image $I$ and an edit instruction $T$, we (i) run a single forward pass of the VLM on $(I, T)$ to extract intermediate hidden states, (ii) feed these states to the trained Q-Former (Section~\ref{supp:data}, Section~\ref{lab:gt-bbox}) to predict a bounding box for the target object, (iii) overlay the predicted box onto $I$ using the same rendering procedure as during training, (iv) encode the overlaid image to the DiT's latent space via the VAE, and (v) run the LoRA-augmented DiT, conditioned on the instruction embedding and the overlaid latent, to produce the edited image. The DiT both localizes its modification to the boxed region and removes the box artifact from the final output.

\subsection{Compute Resources.}
Q-Former training requires approximately 60\,GB of VRAM and completes in roughly 2 hours per run on a single NVIDIA A100 80\,GB GPU.
LoRA fine-tuning of the DiT likewise uses a single A100 80\,GB node and requires approximately 5 hours of training time.

\section{Additional Results}
\label{supp:results}

\subsection{Complete Evaluation Metrics}
\label{supp:evals}
Table \ref{tab:full_metrics} expands on the main text by providing the complete set of evaluation metrics, including individual breakdown scores rather than just the mean VQA score.

The mean VQA score reported in the main paper aggregates three questions, each scored on a $0$--$4$ scale, that target complementary aspects of edit quality:
\begin{itemize}
    \item \textbf{Overall edit instruction adherence.} A holistic judgment of whether the edited image, as a whole, realizes what the instruction asked for. 
    \item \textbf{Subject modification accuracy.} Localized correctness on the target subject identified by the ground-truth bounding box: was \emph{this} object modified in the way the instruction specified? 
    \item \textbf{Background preservation and leakage prevention.} Penalizes leakage by asking whether everything outside the intended edit region was left untouched.
\end{itemize}

Our method achieves the best overall results, leading across all VQA-based evaluations while performing comparably to the strongest baselines in the preservation of the input image.
\input{tables/full_metrics_icml}

\subsection{Additional Qualitative Results}
\label{supp:qualitative}

We provide further qualitative evidence supporting the claims in the main paper. Figure~\ref{fig:localization} contrasts successful Q-Former bounding-box predictions with the corresponding localization failures of the baseline Qwen-Image-Edit pipeline on the same inputs, illustrating how the recovered spatial signal directly addresses the failure mode identified in our analysis. Figure~\ref{fig:qual_results_supp} extends the qualitative comparison of Figure~\ref{fig:qual_results} with two additional samples evaluated against the same set of baselines.

\input{figures/localization_icml}
\input{figures/qualitative_supp_icml}

\section{Broader Impacts}
\label{supp:impacts}

This work advances the understanding of how Vision-Language Models (VLMs) encode spatial information when run as part of an image editing pipeline, and leverages these insights to improve the precision of text-guided image editing. On the positive side, our Analysis-by-Proxy framework contributes to the growing field of VLM interpretability, offering a transparent method to examine the internal mechanisms of multimodal encoders. Practically, providing users with more reliable and precisely localized editing tools lowers the barrier to entry for creative professionals and everyday users.

However, we acknowledge the inherent dual-use risks associated with improvements in generative editing capabilities. Enhancing the spatial accuracy and structural preservation of image editing models makes it easier to seamlessly modify visual content, which could be misused to generate deepfakes, manipulate imagery, or spread disinformation. While our research focuses on the architectural analysis and foundational understanding of these pipelines, the resulting techniques could be exploited maliciously. Mitigating these societal risks will require continued investment in parallel defenses, such as robust watermarking, image provenance standards, and manipulation detection systems. These are areas where deeper architectural interpretability, like the insights provided in this work, may also prove beneficial.

\section{Assets and Licenses}
\label{supp:assets}

In this work, we utilize several existing models, architectures, and evaluation metrics, all of which are properly cited in the main text and used strictly for research purposes in accordance with their respective terms. Specifically, we build upon the open-weights Qwen2.5-VL and Qwen-Image-Edit pipelines~\cite{wu2025qwen} and utilize the Q-Former architecture~\cite{li2023blip2bootstrappinglanguageimagepretraining} for our proxy models. For our baseline comparisons and quantitative metrics, we evaluate against FLUX-Kontext~\cite{labs2025flux1kontextflowmatching} and FLUX.2~\cite{flux-2-2025} , FIBO-Edit~\cite{gutflaish2025generatingimage1000words}, and the LPIPS metric~\cite{zhang2018unreasonableeffectivenessdeepfeatures}. All automated Vision Question Answering evaluations and prompt generations utilizing the Gemini 2.5 Pro API~\cite{gemini25report2025} were conducted in compliance with the Google Cloud Terms of Service. All dataset images were generated by the authoers.

To ensure our findings can be freely reproduced and extended, we will release all new assets introduced in this paper upon publication in order to preserve anonymity. This includes the trained Q-Former proxy checkpoints, the fine-tuned LoRA module for Qwen-Image-Edit, our synthetic spatial training set, and the curated 200-sample evaluation set. All custom code and model weights are released under the MIT License, while the datasets and human-annotated bounding boxes are distributed under the Creative Commons Attribution 4.0 International (CC BY 4.0) License.

\clearpage
\appendix
\section*{\Large Appendix}
\section{Dataset Generation Prompts}
\label{app:dataset_prompts}

To generate the initial scene descriptions and their corresponding edit instructions in Section \ref{supp:data}, we provide the following prompt to the language model. The prompt is designed to yield diverse multi-object scenes in which a single object acts as a deliberate anomaly, establishing the foundation for our editing pipeline. We use the following system prompt:

\begin{tcolorbox}[colback=gray!10, colframe=gray!50, boxrule=0.5pt]
\ttfamily
Create a list of 200 pairs of prompts. The generation prompts should generate diverse and interesting images, each containing 4 to 9 objects of the same base type, organized in random layouts and varied scenes. You may use templates such as:
\begin{itemize}
    \item ``A photo of \{\}''
    \item ``A high-resolution realistic image of \{\}''
    \item ``A close-up photo of a \{\}''
\end{itemize}
This is not an exhaustive list. The goal is to obtain a set of diverse images that can be used to test the local editing capabilities of models.

Each generation prompt must contain $N-1$ identical objects, and one anomalous object that differs clearly in color, shape, or appearance. Each edit prompt must instruct the model to convert the anomalous object to match the other $N-1$ objects exactly.

\textbf{Example:}
\begin{itemize}
    \item \textbf{GENERATION:} An image of 7 dogs in a line; 6 are Labradors and one is a Husky.
    \item \textbf{EDIT:} Turn the Husky into a Labrador, like all the other dogs.
\end{itemize}
\end{tcolorbox}

This structure provides pairs of images that allow us to later ``invert'' the edit: mapping from a uniform set to a single edited anomaly.

The downstream image generation model is capable of text-to-JSON prompt enhancement, enabling precise spatial layouts. An example of an enhanced scene description for this task is:

\begin{quote}
\textit{``Hyper-detailed, ultra-fluffy owls sitting in the trees at night, looking directly at the camera. There are 7 owls in total. Their feathers are soft and voluminous, slightly different colors, catching the cool moonlight with subtle silver highlights. The owls' gaze is curious and full of charm, giving it a whimsical, storybook-like personality.''}
\end{quote}

%% file: tables/full_metrics_icml.tex
\begin{table*}[t]
\label{supp:quant-table}
\centering
\caption{Complete evaluation metrics including all VQA sub-scores. LPIPS and L2 measure background preservation (lower is better). VQA metrics evaluate edit correctness, instruction adherence, and resistance to instruction leakage (higher is better). Values are mean $\pm$ 1\,SEM over $n{=}200$ test samples (sample std with $\text{ddof}{=}1$, divided by $\sqrt{n}$).}
\small
\setlength{\tabcolsep}{4pt}
\begin{tabular}{lccccc}
\toprule
Method & LPIPS comp. $\downarrow$ & L2 comp. $\downarrow$ & VQA edit $\uparrow$ & VQA accuracy $\uparrow$ & VQA leakage $\uparrow$ \\
\midrule
Ours & \underline{0.0794}{\tiny$\,\pm\,$0.0046} & \underline{0.0070}{\tiny$\,\pm\,$0.0010} & \textbf{3.2550}{\tiny$\,\pm\,$0.0964} & \textbf{3.3300}{\tiny$\,\pm\,$0.0916} & \textbf{3.5500}{\tiny$\,\pm\,$0.0838} \\
Qwen Image Edit & 0.1191{\tiny$\,\pm\,$0.0053} & 0.0122{\tiny$\,\pm\,$0.0012} & 3.1450{\tiny$\,\pm\,$0.0960} & \underline{3.2250}{\tiny$\,\pm\,$0.0988} & 3.1450{\tiny$\,\pm\,$0.1113} \\
Fibo edit & 0.2560{\tiny$\,\pm\,$0.0084} & 0.0181{\tiny$\,\pm\,$0.0013} & 2.5100{\tiny$\,\pm\,$0.1270} & 2.4600{\tiny$\,\pm\,$0.1295} & 2.1200{\tiny$\,\pm\,$0.1241} \\
FLUX.2 dev & 0.1377{\tiny$\,\pm\,$0.0081} & 0.0172{\tiny$\,\pm\,$0.0021} & 2.6100{\tiny$\,\pm\,$0.1129} & 2.6650{\tiny$\,\pm\,$0.1148} & 2.3600{\tiny$\,\pm\,$0.1336} \\
Flux kontext & 0.1847{\tiny$\,\pm\,$0.0123} & 0.0248{\tiny$\,\pm\,$0.0023} & 2.1050{\tiny$\,\pm\,$0.1164} & 2.5850{\tiny$\,\pm\,$0.1170} & 2.0850{\tiny$\,\pm\,$0.1374} \\
Full description & 0.0818{\tiny$\,\pm\,$0.0059} & 0.0080{\tiny$\,\pm\,$0.0012} & \underline{3.2050}{\tiny$\,\pm\,$0.0991} & 3.2000{\tiny$\,\pm\,$0.0975} & 3.4150{\tiny$\,\pm\,$0.0911} \\
Norm-AVG & \textbf{0.0700}{\tiny$\,\pm\,$0.0037} & \textbf{0.0065}{\tiny$\,\pm\,$0.0009} & 2.8050{\tiny$\,\pm\,$0.1168} & 2.6700{\tiny$\,\pm\,$0.1184} & \underline{3.4250}{\tiny$\,\pm\,$0.0947} \\
\bottomrule
\end{tabular}
\label{tab:full_metrics}
\end{table*}

%% file: figures/localization_icml.tex
\begin{figure}[t]
    \centering
    \setlength{\tabcolsep}{2pt}
    \begin{tabular}{@{}c@{\hspace{2pt}}c@{\hspace{4pt}}c@{}}
        & \parbox[t]{0.30\linewidth}{\centering \small ``Recolour the bottom right clownfish to be black and white''}
        & \parbox[t]{0.30\linewidth}{\centering \small ``Make the bottom jellyfish glow red''}
        \\[6pt]

        \parbox[c][0.30\linewidth][c]{2em}{\centering \rotatebox[origin=c]{90}{\small \textbf{\begin{tabular}{@{}c@{}}Successful \\ localization\end{tabular}}}} &
\raisebox{-0.5\height}{\includegraphics[width=0.30\linewidth]{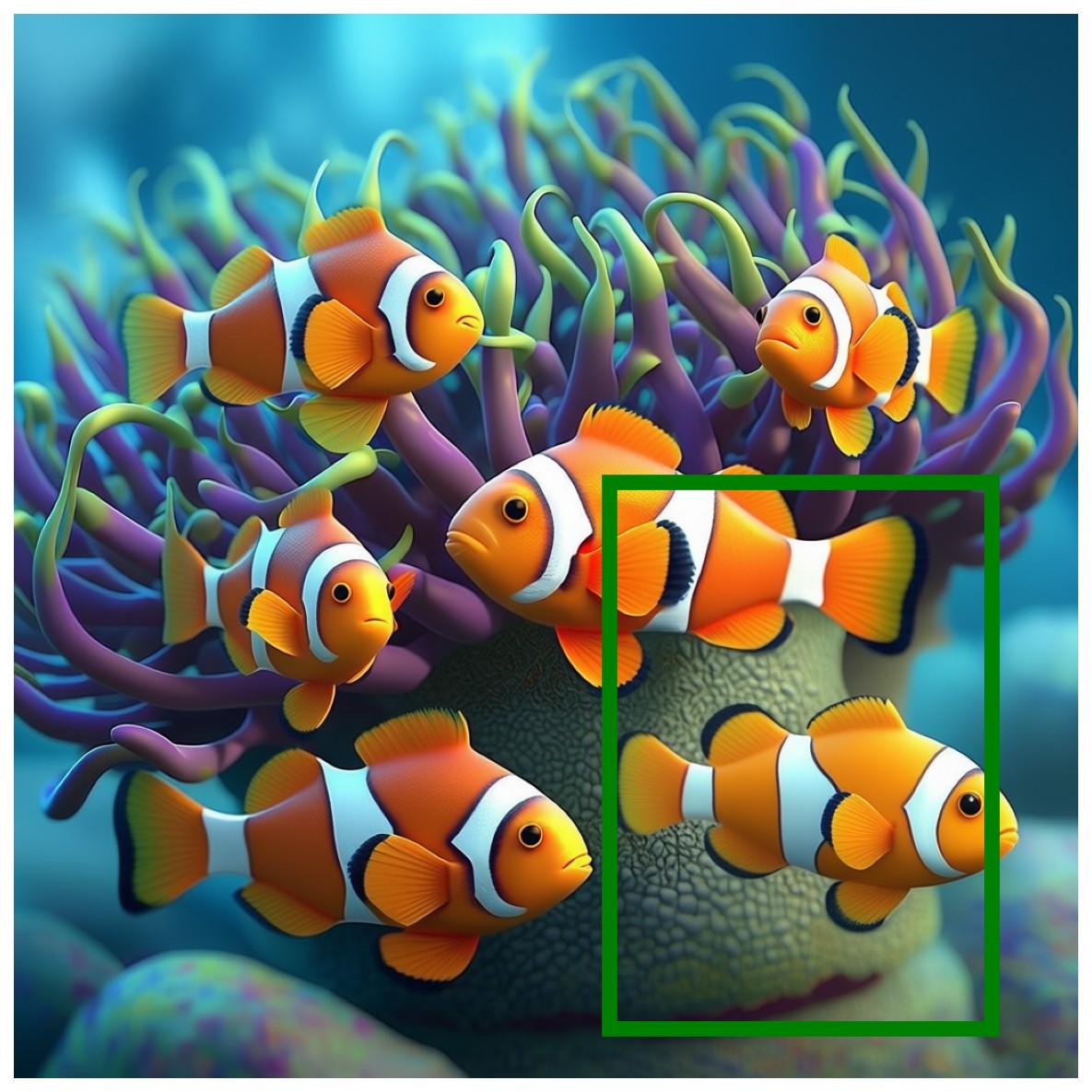}} &
\raisebox{-0.5\height}{\includegraphics[width=0.30\linewidth]{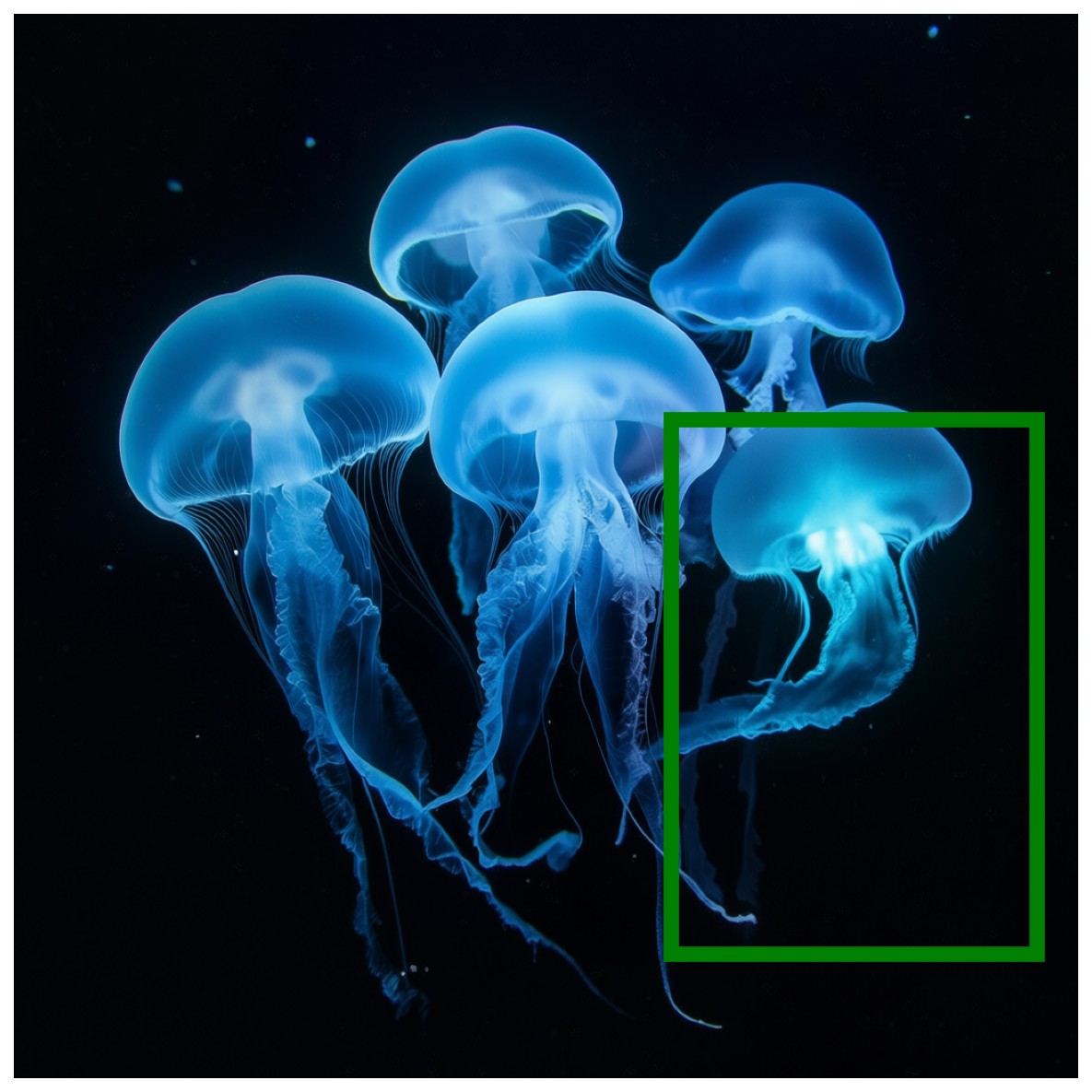}} \\

        \parbox[c][0.30\linewidth][c]{2em}{\centering \rotatebox[origin=c]{90}{\small \textbf{Failed Edit}}} &
        \raisebox{-0.5\height}{\includegraphics[width=0.30\linewidth]{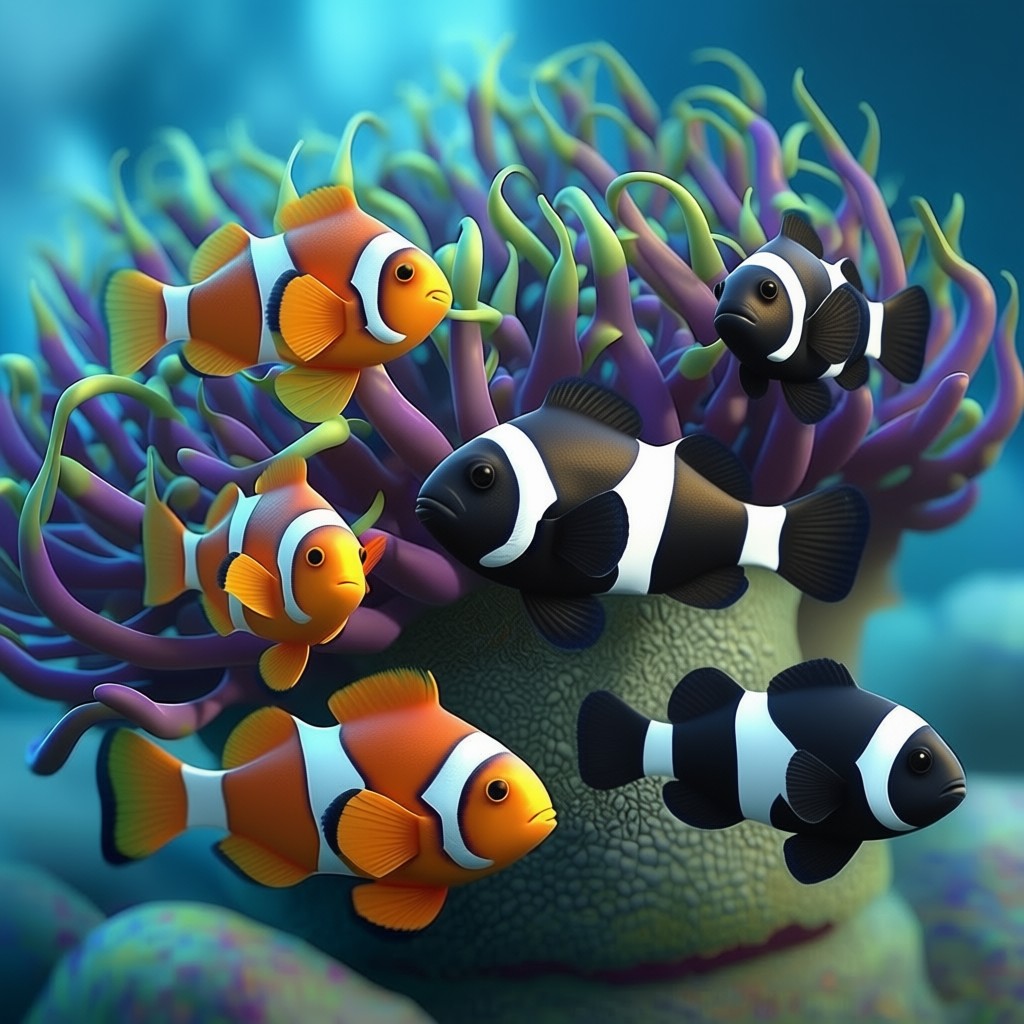}} &
        \raisebox{-0.5\height}{\includegraphics[width=0.30\linewidth]{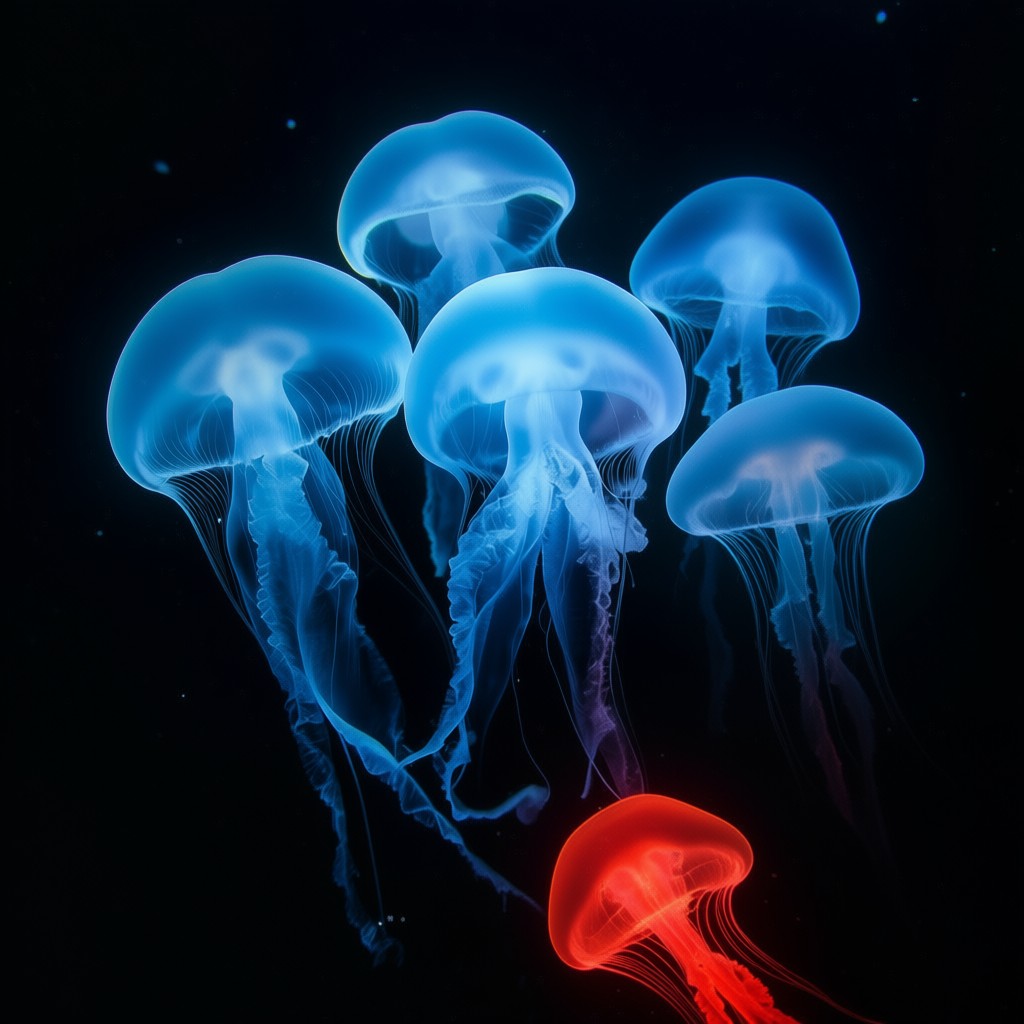}} \\
    \end{tabular}
    \caption{\textcolor{green!50!black}{\textbf{Green}}: the explicit bounding box prediction generated by the trained Q-Former proxy, which isolates the target region to guide the downstream localized edit. The bottom row shows the corresponding failures of the original Qwen-Image-Edit pipeline on the same inputs.}
    \label{fig:localization}
\end{figure}

%% file: figures/qualitative_supp_icml.tex
\begin{figure*}[t]
    \centering
    \begin{minipage}{0.95\linewidth}
    \centering
    \scriptsize
    \setlength{\tabcolsep}{1.5pt} %

    \begin{tabular}{c c c c c c c c}
        \tiny \textbf{Input} & \tiny \textbf{Ours} & \tiny \textbf{Baseline} & \tiny \textbf{Kontext} & \tiny \textbf{FIBO} & \tiny \textbf{FLUX.2} & \tiny \textbf{Full desc} & \tiny \textbf{Norm-AVG} \\[6pt]

        \multicolumn{8}{c}{\tiny ``Change the black sand in the leftmost hourglass to pure white sand.''} \\[2pt]
        \includegraphics[width=0.115\linewidth]{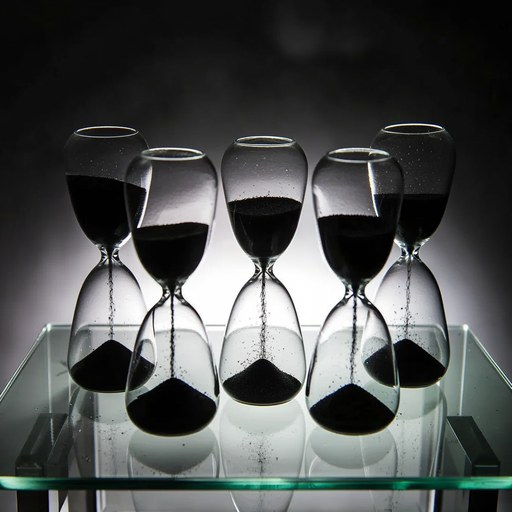} &
        \includegraphics[width=0.115\linewidth]{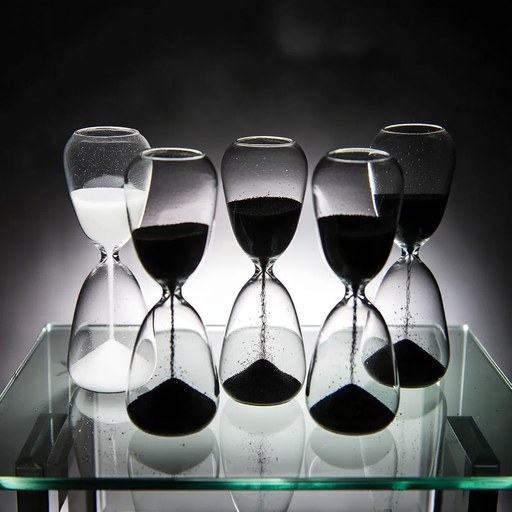} &
        \includegraphics[width=0.115\linewidth]{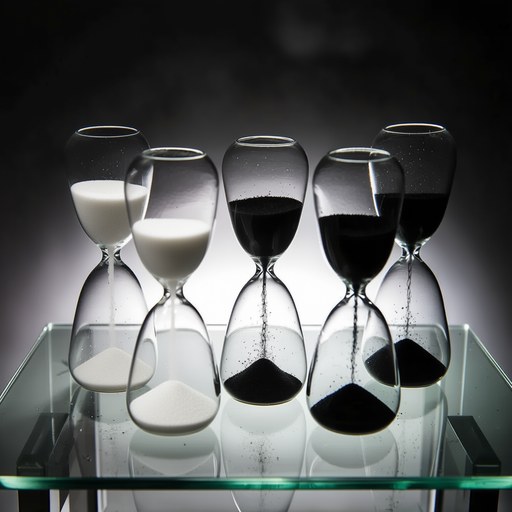} &
        \includegraphics[width=0.115\linewidth]{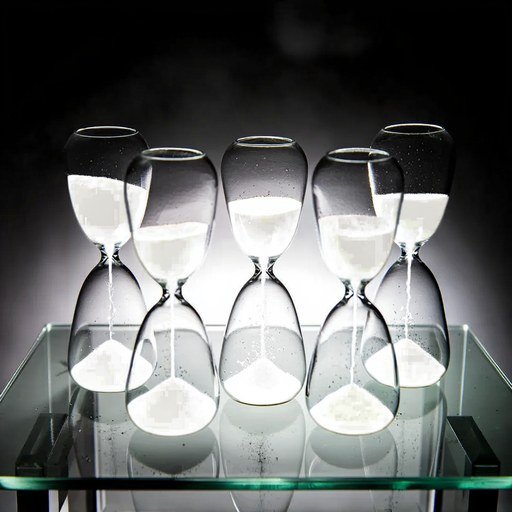} &
        \includegraphics[width=0.115\linewidth]{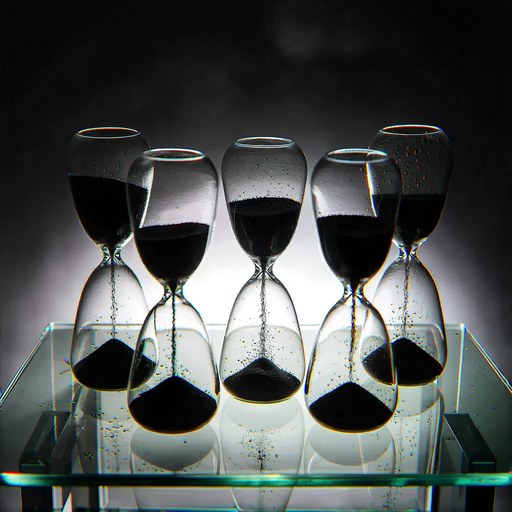} &
        \includegraphics[width=0.115\linewidth]{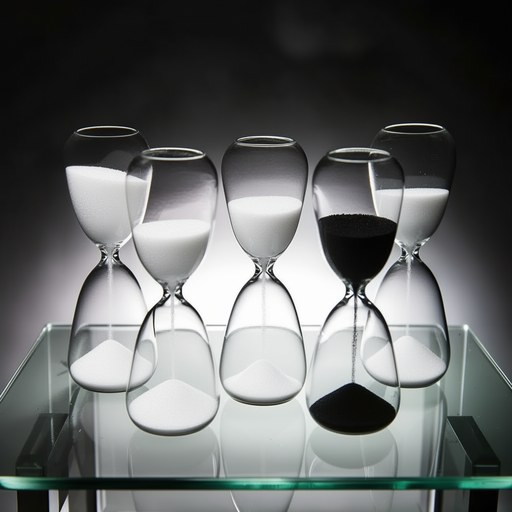} &
        \includegraphics[width=0.115\linewidth]{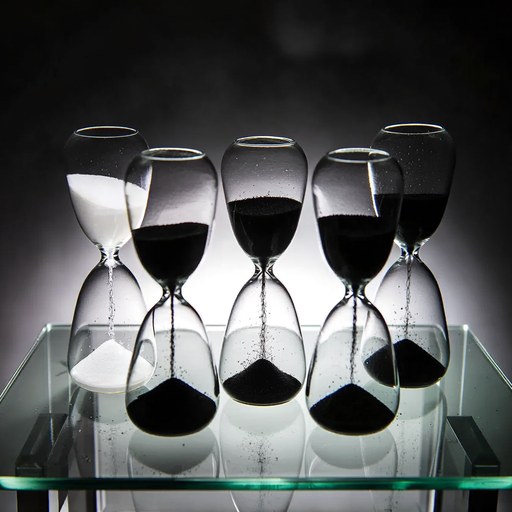} &
        \includegraphics[width=0.115\linewidth]{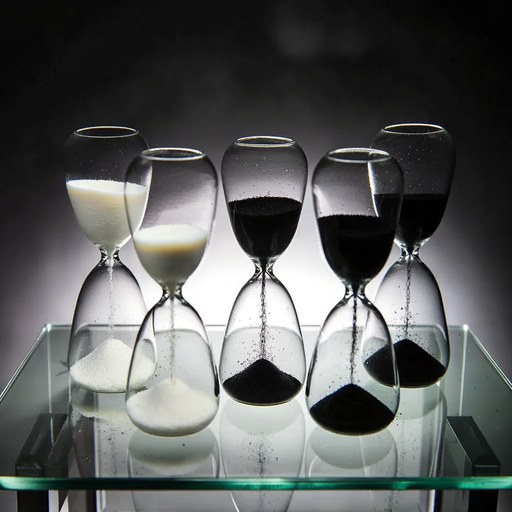} \\[3pt]

        \multicolumn{8}{c}{\tiny ``Change the blinking LED lights on the leftmost server rack to bright blue.''} \\[2pt]
        \includegraphics[width=0.115\linewidth]{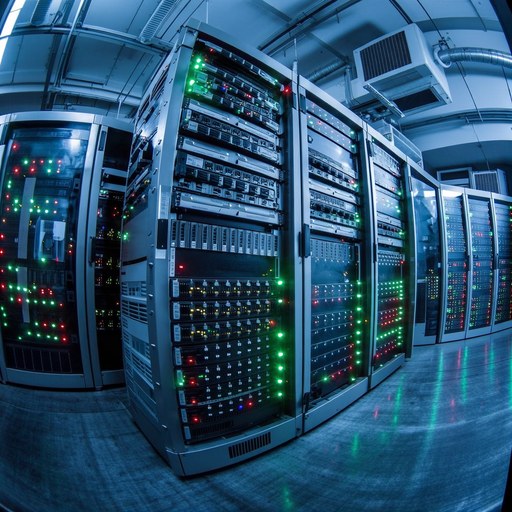} &
        \includegraphics[width=0.115\linewidth]{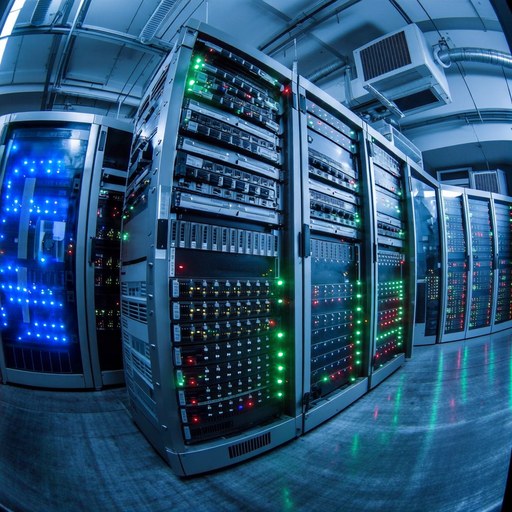} &
        \includegraphics[width=0.115\linewidth]{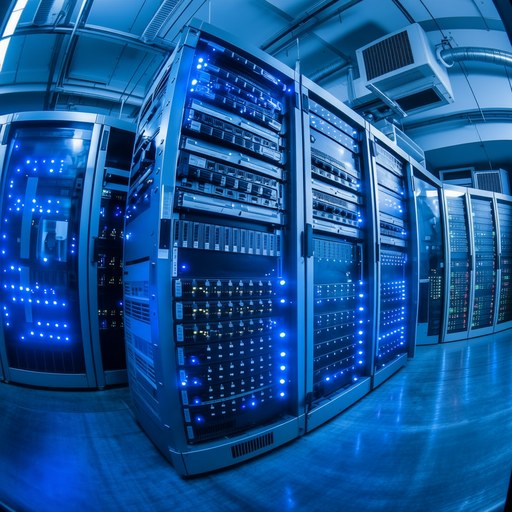} &
        \includegraphics[width=0.115\linewidth]{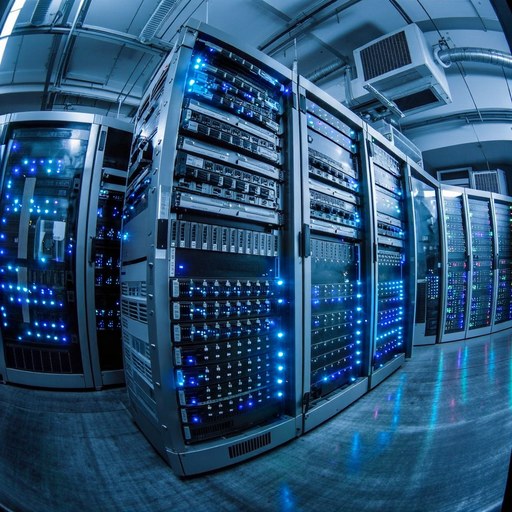} &
        \includegraphics[width=0.115\linewidth]{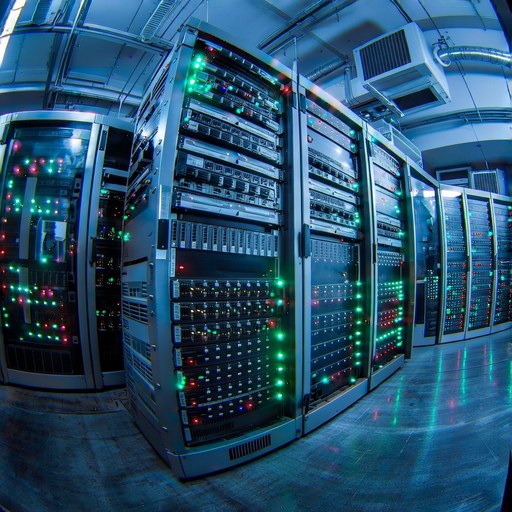} &
        \includegraphics[width=0.115\linewidth]{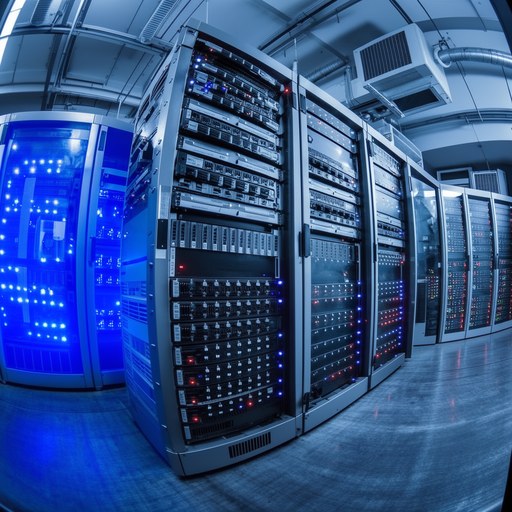} &
        \includegraphics[width=0.115\linewidth]{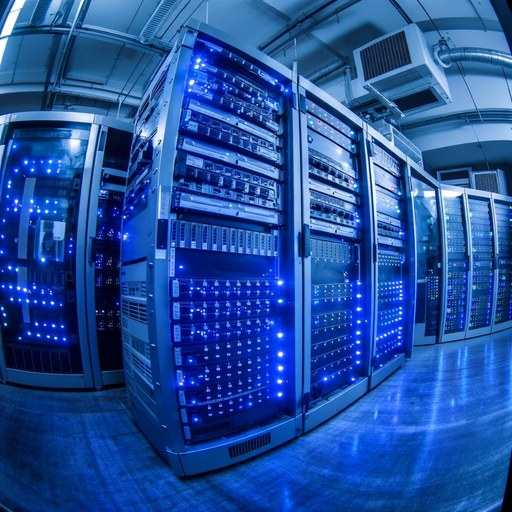} &
        \includegraphics[width=0.115\linewidth]{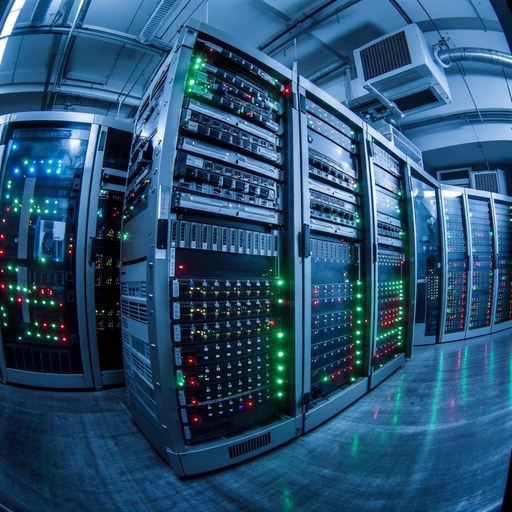}

    \end{tabular}
    \end{minipage}

    \caption{\small Additional qualitative comparisons of image editing results. As in the main paper, our method successfully localizes the edit to the intended target (the lowest quartz cluster, the leftmost server rack), while existing methods frequently modify incorrect objects or apply changes broadly across the scene.}
    \label{fig:qual_results_supp}
\end{figure*}